\documentclass{article} %
\usepackage[preprint]{iclr2027_conference}
\usepackage{times}
\usepackage[OT1]{fontenc}

\usepackage{amsmath,amsfonts,bm}

\def\eqref#1{equation~\ref{#1}}

\def\1{\bm{1}}

\DeclareMathAlphabet{\mathsfit}{\encodingdefault}{\sfdefault}{m}{sl}
\SetMathAlphabet{\mathsfit}{bold}{\encodingdefault}{\sfdefault}{bx}{n}

\usepackage{hyperref}
\usepackage{url}

\usepackage{todonotes}
\usepackage{graphicx}   %
\usepackage{wrapfig}    %
\usepackage{booktabs}   %
\usepackage{amsmath}
\usepackage{amssymb}
\usepackage{xspace}     %
\usepackage[nameinlink,capitalize]{cleveref}
\usepackage{tcolorbox}
\tcbuselibrary{breakable}
\newtcolorbox{vlmprompt}[1]{
  breakable,
  colback=white,
  colframe=black!20,
  colbacktitle=black!6,
  coltitle=black,
  fonttitle=\small\bfseries,
  fontupper=\small\color{black},
  boxrule=0.4pt,
  arc=1mm,
  left=6pt,right=6pt,top=5pt,bottom=5pt,
  before skip=8pt,after skip=8pt,
  title={#1}
}

\definecolor{cvprblue}{rgb}{0.21,0.49,0.74}
\hypersetup{colorlinks,urlcolor=black,citecolor=cvprblue}

\title{Squaring-the-Sphere: Distortion-Free and Seamless  Panorama Generation via CubeAtlas}
\author{\textbf{Xuhui Zuo$^{*}$, Chenjie Cao$^{*}$, Tengfei Wang, Junta Wu, Yisu Zhang, Chunchao Guo$^{\dagger}$}\\
\normalfont Tencent \quad {\footnotesize $^{*}$Equal contribution. \quad $^{\dagger}$Corresponding author.}
}

\newcommand{\methodname}{\textsc{Squaring-the-Sphere}\xspace}

\begin{document}

\maketitle

\begin{abstract}

Generating $360^\circ$ panoramas requires both local perspective fidelity and global spherical continuity.
Existing representations face a trade-off: equirectangular projection (ERP) enables seamless panorama modeling but suffers from severe distortion and a domain gap to pre-trained image generators, whereas cubemaps preserve perspective geometry but often produce inconsistent faces and visible seams.
We present \methodname, a panoramic image generation framework built on a novel representation \textbf{CubeAtlas}, which unfolds all cubemap faces into a single compact canvas.
By arranging orientation-aligned, rotated polar regions alongside the equatorial faces, this representation preserves the perspective geometry of cubemaps while organizing the cross-face relationships required for seamless panorama assembly.
It further enables a pretrained diffusion transformer to model the complete panorama jointly in a single pass, without altering the backbone architecture or positional encoding.
While CubeAtlas has a token count comparable to ERP, its structured layout makes many dense attention interactions unnecessary.
We therefore introduce a layout-aware sparse attention mask that preserves the interactions required for panorama continuity while removing geometrically irrelevant token pairs.
As a result, CubeAtlas achieves better quality than ERP modeling with lower attention computation.
Furthermore, we introduce a training-free super-resolution technique and a fusion scheme that enhance fine details in distortion-free perspective space before achieving the final ERP panorama. 
Experiments on various benchmarks demonstrate that our method improves geometric fidelity, cross-view consistency, and high-resolution visual quality over previous methods.
\end{abstract}

\section{Introduction}
\label{sec:intro}

\begin{figure}[t]
\begin{center}
\includegraphics[width=\linewidth]{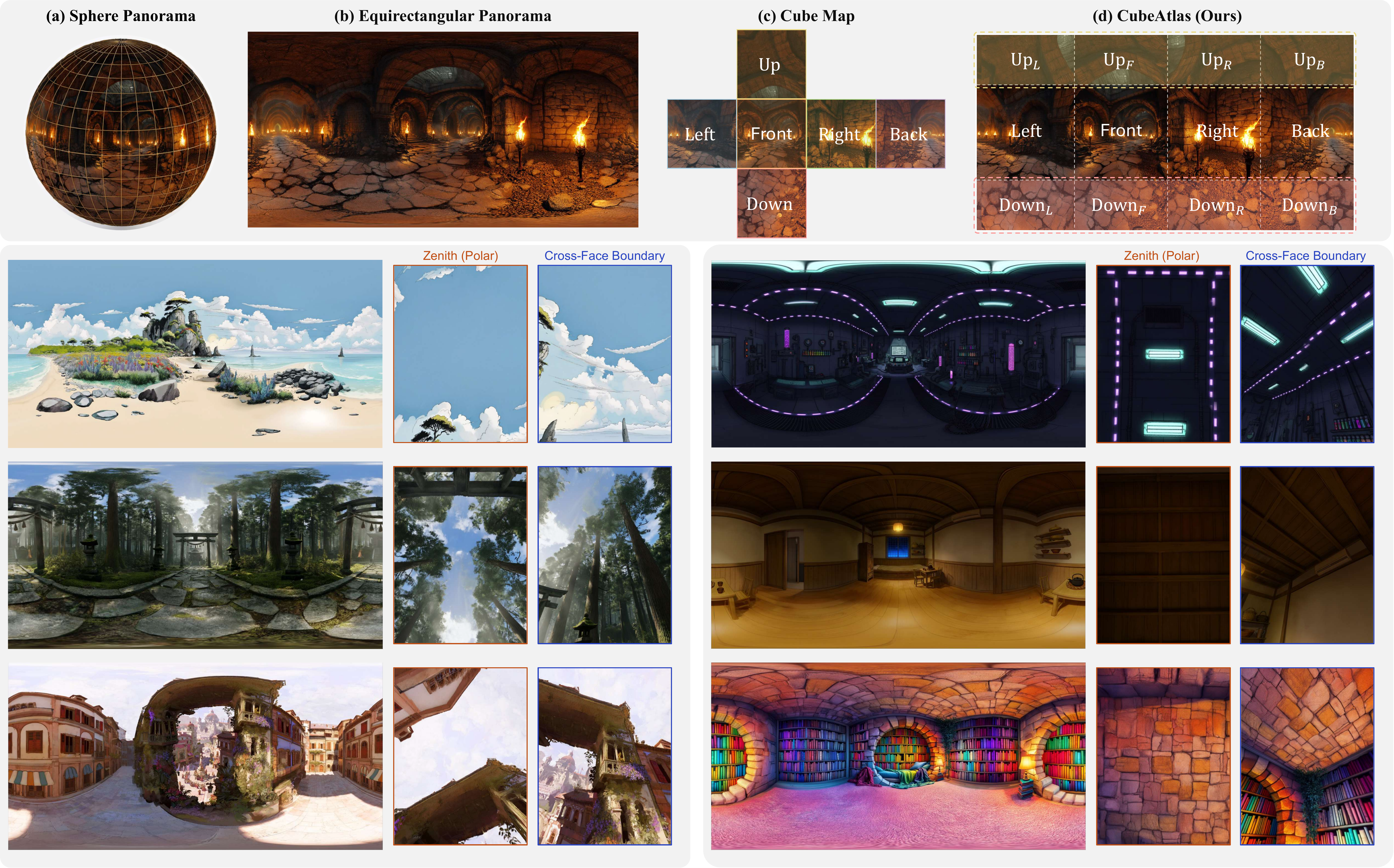}
\end{center}
\vspace{-0.15in}
\caption{\textbf{\methodname reconciles global panorama continuity with
perspective-face geometry.} \emph{Upper:} ERP represents the full panorama on a
single canvas but severely stretches the polar regions; cubemaps preserve
perspective geometry but introduce disconnected face boundaries. CubeAtlas packs the four equatorial faces and rotated half-polar regions into a compact canvas, making their geometric adjacencies local for joint generation. \emph{Lower:} \methodname produces seamless, geometrically coherent panoramas across diverse indoor and outdoor scenes, shown as ERP renderings with perspective crops.\label{fig:teaser}}
\vspace{-0.15in}
\end{figure}

$360^\circ$ panoramas serve as a fundamental visual medium for a broad range of immersive applications, including virtual and augmented reality, gaming, telepresence, and 3D world modeling~\citep{hunyuanworld1,hunyuanworld2,worldlabs_marble,diffpanopp,panoworld}.
Recent advances in image generative models have substantially improved the realism, diversity, and controllability of perspective image synthesis~\citep{esser2024sd3,blackforest2024flux,qwenedit,cao2025hunyuanimage3}.
However, how to transfer the powerful generative priors learned in perspective space to panorama generation remains an open challenge, as a panorama must simultaneously preserve local perspective fidelity and maintain globally coherent, seamless content over the entire sphere (\Cref{fig:teaser}).

In this paper, we identify the key challenge as selecting a panoramic representation compatible with pre-trained image priors.
The two most common choices are equirectangular projection (ERP) and cubemaps.
ERP represents the scene on a single equirectangular canvas, facilitating joint modeling of the entire panorama~\citep{panfusion,diffusion360,anything360,omnix}.
However, its non-uniform spherical projection severely stretches polar regions, introduces geometric distortion, and departs from the perspective-image distribution learned by modern generators~\citep{worldshaper}.
Cubemaps avoid ERP's polar stretching by decomposing the sphere into six perspective faces~\citep{mvdiffusion,cubediff}. Although methods such as CubeDiff jointly model these faces through cross-face attention, preserving spatial continuity across face boundaries remains challenging.
Therefore, unifying ERP's global coherence with cubemaps' perspective fidelity remains a central problem for high-quality panorama generation.

To address this trade-off, we propose \methodname, a framework built on a new panorama representation termed \textbf{CubeAtlas}.
Rather than generating an ERP image or six independent cube faces, \methodname unfolds the complete cubemaps into a single image-like grid and enables a diffusion transformer (DiT)~\citep{peebles2023dit} to model the entire panorama jointly in one forward pass, as illustrated by the CubeAtlas layout in the upper part of \Cref{fig:teaser}.
By combining half-polar rotation and packing for seam-aware yet compact face alignment with layout-aware sparse attention for efficient token interactions, \methodname promotes cross-view continuity without the cost of dense global attention.
Crucially, our representation is compatible with existing DiT-based image generators without modifying their backbone architectures or positional encodings, thereby inheriting large-scale 2D image priors with minimal adaptation.

\begin{figure*}[t]
\centering
\includegraphics[width=0.8\textwidth]{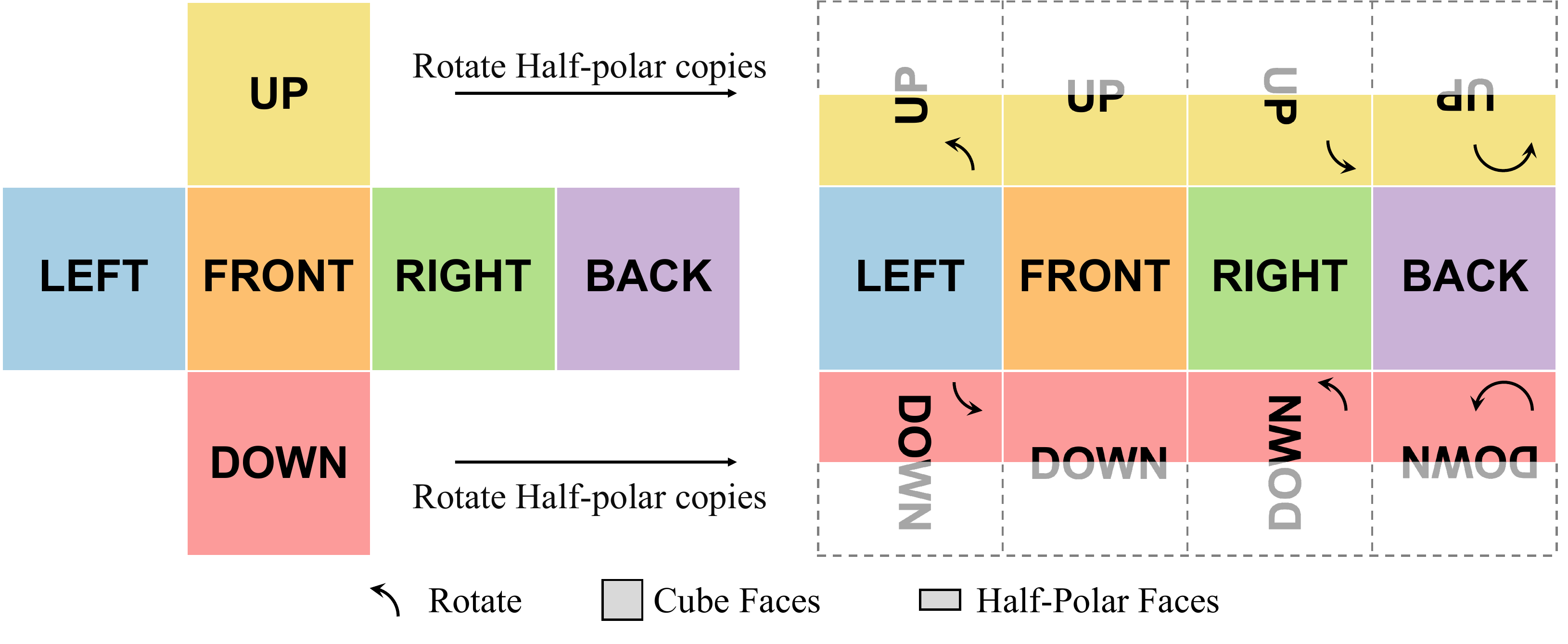}
\vspace{-0.15in}
\caption{\textbf{CubeAtlas representation.} Four equatorial faces are unfolded into the central band. The upper and lower faces are rotated into orientation-aligned half-polar copies and placed next to their corresponding equatorial faces, forming a compact layout for joint generation.\label{fig:cubeatlas}}
\vspace{-0.15in}
\end{figure*}

Concretely, CubeAtlas re-arranges the six cubemap faces into a compact three-band, four-column layout, as detailed in \Cref{fig:cubeatlas}. For each equatorial orientation, it retains only the equator-adjacent half of the upper and lower faces, rotated to align with the corresponding equatorial view.
This half-polar layout retains the local context needed to connect each polar region to its neighboring equatorial faces while avoiding redundant polar tokens; complementary rotated copies are later combined to reconstruct the complete polar faces.
Polar rotation aligns polar content with the corresponding equatorial views, turning otherwise discontinuous cross-face relationships into locally organized neighborhoods within the grid, thereby facilitating seamless grid-to-ERP assembly.
We further introduce layout-aware sparse attention that preserves the intra-face and continuity-related interactions required for panorama generation while masking irrelevant token pairs. 
Specifically, full attention is retained within each latitude band for horizontal consistency, while cross-latitude communication is mediated solely by the equatorial band: upper and lower regions never interact directly, and each equatorial region attends only to the polar regions on the same left--right or front--back axis.
Finally, we reproject the ERP into overlapping perspective cube views and apply a training-free $4\times$ super-resolution model~\citep{lu2026pid} to each view. Low-frequency correction preserves coherent color and illumination in the final ERP, while linear blending of overlaps solves discontinuities.

We summarize three contributions of this paper as follows:
\begin{itemize}
    \item \textbf{CubeAtlas representation} is proposed that unfolds an entire cubemap into a single compact canvas with orientation-aligned half-polar regions. It retains only the equator-adjacent portions of rotated polar copies to reduce redundant polar tokens while preserving the context required for full polar reconstruction. The representation preserves the perspective geometry of cubemaps, while resolving cross-face discontinuities, enabling seamless panorama generation and direct reuse of 2D image priors from pre-trained DiT generators.

    \item \textbf{Layout-aware sparse attention} retains full attention within each latitude band while restricting cross-band interactions according to the spatial structure of CubeAtlas. This structured connectivity preserves the interactions needed for joint panorama modeling while avoiding unnecessary all-to-all attention among target tokens.

    \item \textbf{Comprehensive evaluation.} We evaluate panorama generation from
    complementary perspectives, including distributional fidelity, ERP and cubemap
    boundary continuity, equatorial and polar perceptual quality, geometric
    distortion, text adherence, and human preference. Extensive comparisons and
    controlled ablations across diverse benchmarks demonstrate consistent gains in
    cross-face continuity and spherical geometric fidelity.
\end{itemize}

\section{Related Work}
\label{sec:related}

\noindent\textbf{360$^{\circ}$ Panorama Generation.}
Existing panorama generators are largely distinguished by how they represent the
sphere~\citep{asig}. 
\textit{ERP-based methods} synthesize a single canvas, which facilitates global generation~\citep{text2light,panfusion,diffusion360}, but polar stretching creates a mismatch with the perspective-image distribution of pretrained generators. 
Panorama-specific adaptations address this mismatch through seam-aware processing~\citep{anything360,sherpa,gimbal360} or hybrid perspective--panorama training~\citep{dit360}.
\textit{Cubemap-based methods} instead preserve perspective geometry while explicitly adapting the model to communicate across separate faces.
CubeDiff~\citep{cubediff} jointly denoises six faces through inflated attention and augments their latents with spherical UV coordinates, while DreamCube~\citep{dreamcube} uses spherical coordinates and synchronizes attention to improve boundary consistency.
Although both support joint generation within a single forward pass, they retain separate face-wise spatial grids and introduce panorama-specific positional signals or operators to mitigate cross-face seams. 
\textit{Our method} instead packs orientation-aligned cubemap faces into a single CubeAtlas canvas, turning cross-face adjacencies into local 2D neighborhoods and enabling seam-consistent panorama generation with the pretrained DiT's native positional encoding and backbone structure.

\noindent\textbf{Image-to-Panorama Generation.}
Image-to-panorama methods extend a narrow-field perspective image into a complete
environment. Earlier diffusion-based approaches typically formulate the task as
panoramic outpainting, using camera-aware projection and inpainting to align the
conditioning image with an ERP target~\citep{panodiffusion}. More recently,
360Anything~\citep{anything360} and HY-Pano~2.0~\citep{hunyuanworld2} 
learn the perspective-to-ERP mapping directly from data in a geometry-free
manner, without explicit camera metadata or camera calibration. \methodname shares this geometry-free
formulation: it consumes the conditioning image as-is, without camera calibration,
explicit input projection, or geometric inpainting. Unlike methods that directly
generate ERP, it learns this mapping in the CubeAtlas, whose perspective-aligned
layout is better matched to the target image prior.

\noindent\textbf{Priors from Generative Image Models.}
Large-scale diffusion and flow-based image generators~\citep{esser2024sd3,blackforest2024flux,qwenedit} provide strong priors for high-fidelity, controllable perspective-image synthesis. Directly
transferring these planar priors to ERP panoramas is nevertheless non-trivial:
their standard spatial representations do not encode spherical periodicity or
polar convergence. Recent training-free adaptations, e.g., SpheRoPE~\citep{sphererope} and SphereDiff~\citep{spherediff}, modify
positional encoding or use spherical projection at inference time to inject these spherical constraints into frozen generators~\citep{lfmultidiffusion}. Such methods demonstrate
the useful panoramic priors already present in foundation models, but their
generation capability must still be induced at inference time rather than learned in a
panorama-specific representation. \methodname instead fine-tunes a pre-trained
generative DiT on CubeAtlas formulations, preserving its perspective-image prior while
learning joint cross-face generation in a representation designed for panoramic
continuity.

\section{Method}
\label{sec:method}

\begin{figure*}[t]
\centering
\includegraphics[width=\textwidth]{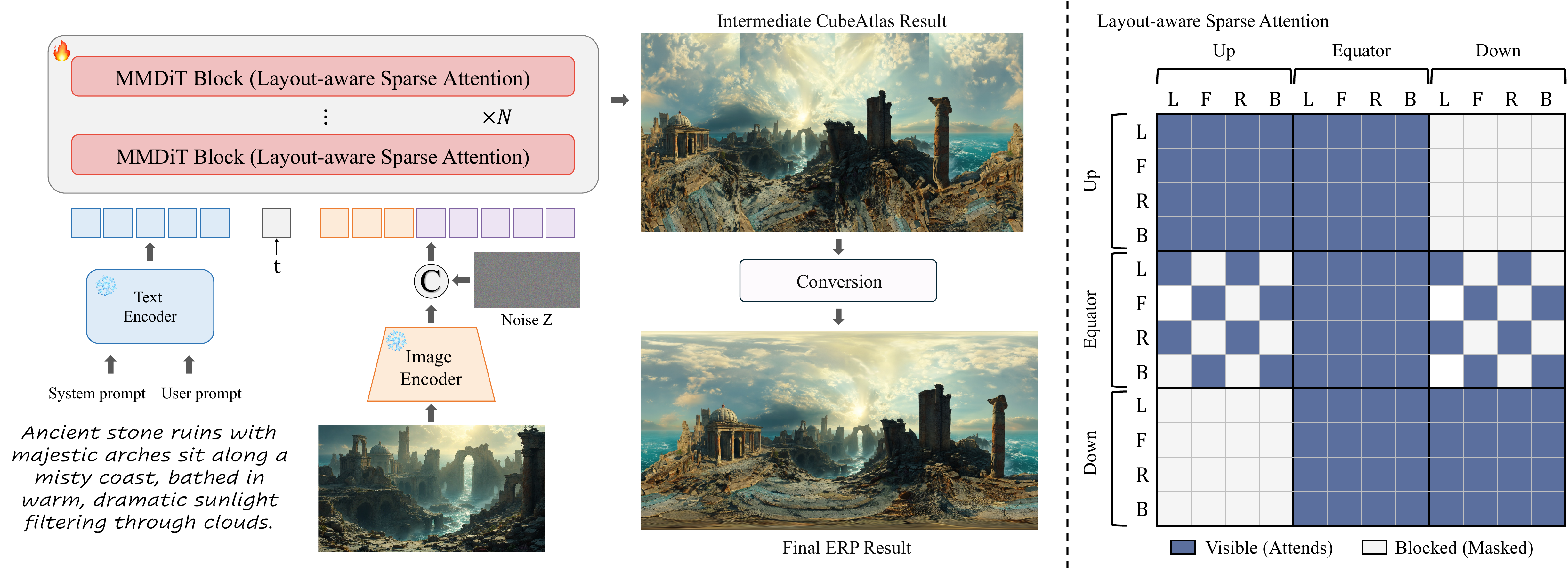}
\vspace{-0.15in}
\caption{\textbf{Overview of \methodname.} Given a perspective image and text
prompts, the DiT denoises the complete CubeAtlas in a single pass.
Layout-aware sparse attention retains within-band interactions while restricting cross-latitude communication: for each pole, an equatorial region attends only to the two axis-aligned half-polar regions that reconstruct a complete polar face.
The generated CubeAtlas is then fused and stitched into the final ERP panorama.\label{fig:pipeline}}
\vspace{-0.15in}
\end{figure*}

We introduce \methodname, an effective approach for generating
high-quality, seamless $360^{\circ}$ panoramas, as provided in
\Cref{fig:pipeline}. 
We first motivate the CubeAtlas layout by reviewing common
panorama representations in \Cref{sec:grid}. 
Then we describe model details in \Cref{sec:train}. 

\subsection{Panoramic Representation}
\label{sec:grid}
A panorama captures a complete $360^{\circ}\times180^{\circ}$ view of a scene from a fixed viewpoint, and the choice of how to lay this view out on a flat image plane has a decisive effect on generation quality. 

\subsubsection{Previous Representations}

\noindent\textbf{Equirectangular Projection (ERP).}
ERP maps longitude and latitude onto the horizontal and vertical axes of a $2:1$ rectangular image, providing a simple and compact representation of the full sphere. 
However, its sampling becomes increasingly non-uniform toward the poles, causing severe horizontal stretching and departing from the perspective-image domain of pretrained diffusion models. 
This mismatch makes it difficult to generate faithful geometry in polar regions. Moreover, spherical continuity is split across the left and right image boundaries, requiring explicit handling of horizontal wrap-around.

\noindent\textbf{Cubemap.}
A cubemap projects the sphere onto six perspective faces, each with a $90^{\circ}$ Field-of-View (FoV). 
It avoids the severe polar stretching of ERP and better aligns with the perspective-image domain of pretrained diffusion models.
However, content that is continuous on the sphere is split across separate face boundaries. 
Without effective cross-face modeling, generated faces can become inconsistent in appearance, semantics, and geometry, resulting in visible seams along cube edges.

\subsubsection{CubeAtlas}

To unify both advantages, we propose a new representation that retains the perspective-face geometry of cubemaps while organizing their cross-face relationships into a single image-like canvas. 
Let $S$ denote the side length of each cube face in latent space. We denote the six standard cubemap orientations by $\Omega=\{\mathrm{L},\mathrm{F},\mathrm{R},\mathrm{B},\mathrm{U},\mathrm{D}\}$, corresponding to left, front, right, back, up, and down. The four equatorial orientations ($\mathrm{L}, \mathrm{F}, \mathrm{R}, \mathrm{B}$) are placed side by side in the central band. 
Jointly denoising this complete band in one forward pass allows the model to learn the horizontal continuity directly.

The more challenging boundaries are vertical. 
In a conventional cubemap-cross representation, only the front face is directly adjacent to the upper and lower faces; the left, right, and back faces are spatially separated from their corresponding polar neighborhoods and receive incompatible positional relationships. 
CubeAtlas therefore creates an in-plane-rotated copy of each polar face for every equatorial orientation, such that its equator-adjacent half is vertically aligned with the corresponding equatorial face. 
We denote these upper half-regions by $\mathrm{U}_{\mathrm{L}},\mathrm{U}_{\mathrm{F}},\mathrm{U}_{\mathrm{R}},\mathrm{U}_{\mathrm{B}}$, while the lower ones are $\mathrm{D}_{\mathrm{L}},\mathrm{D}_{\mathrm{F}},\mathrm{D}_{\mathrm{R}},\mathrm{D}_{\mathrm{B}}$.
The core layout of CubeAtlas can be presented as:
\begin{equation}
\label{eq:cubeatlas-layout}
\mathrm{CubeAtlas} =
\begin{bmatrix}
\mathrm{U}_{\mathrm{L}} & \mathrm{U}_{\mathrm{F}} & \mathrm{U}_{\mathrm{R}} & \mathrm{U}_{\mathrm{B}}\\
\mathrm{L} & \mathrm{F} & \mathrm{R} & \mathrm{B}\\
\mathrm{D}_{\mathrm{L}} & \mathrm{D}_{\mathrm{F}} & \mathrm{D}_{\mathrm{R}} & \mathrm{D}_{\mathrm{B}}
\end{bmatrix}
\in \mathbb{R}^{2S\times4S\times C},
\end{equation}
where $C$ is the number of latent channels. The equatorial band $[\mathrm{L},\mathrm{F},\mathrm{R},\mathrm{B}]$ has a shape $S\times4S\times C$, while the upper and lower polar bands $[\mathrm{U}_{\mathrm{L}},\mathrm{U}_{\mathrm{F}},\mathrm{U}_{\mathrm{R}},\mathrm{U}_{\mathrm{B}}]$ and $[\mathrm{D}_{\mathrm{L}},\mathrm{D}_{\mathrm{F}},\mathrm{D}_{\mathrm{R}},\mathrm{D}_{\mathrm{B}}]$ each has a shape $(S/2)\times4S\times C$.
Thus, each equatorial face is locally connected to its corresponding upper and lower neighborhoods. 
Meanwhile, full attention within each polar band lets the differently oriented half-regions
learn a consistent polar appearance, jointly resolving the cubemap seams.

Although each pole is represented by four rotated copies, CubeAtlas retains only the equator-adjacent half of each copy. 
Its core canvas therefore has size $2S\times4S\times C$, equal to an ERP latent canvas at the same resolution, while retaining the perspective geometry of cubemap faces. This half-polar packing avoids redundant polar tokens and keeps the base computation comparable to ERP even before the further savings from our sparse attention mask. %
To encode horizontal periodicity, we append a context strip copied from the left edge to the right boundary before VAE encoding. The resulting CubeAtlas is processed as a single image in one VAE forward pass, producing the latent representation used for diffusion training.

\begin{wrapfigure}{r}{0.33\columnwidth}
\centering
\vspace{-0.2in}
\includegraphics[width=0.8\linewidth]{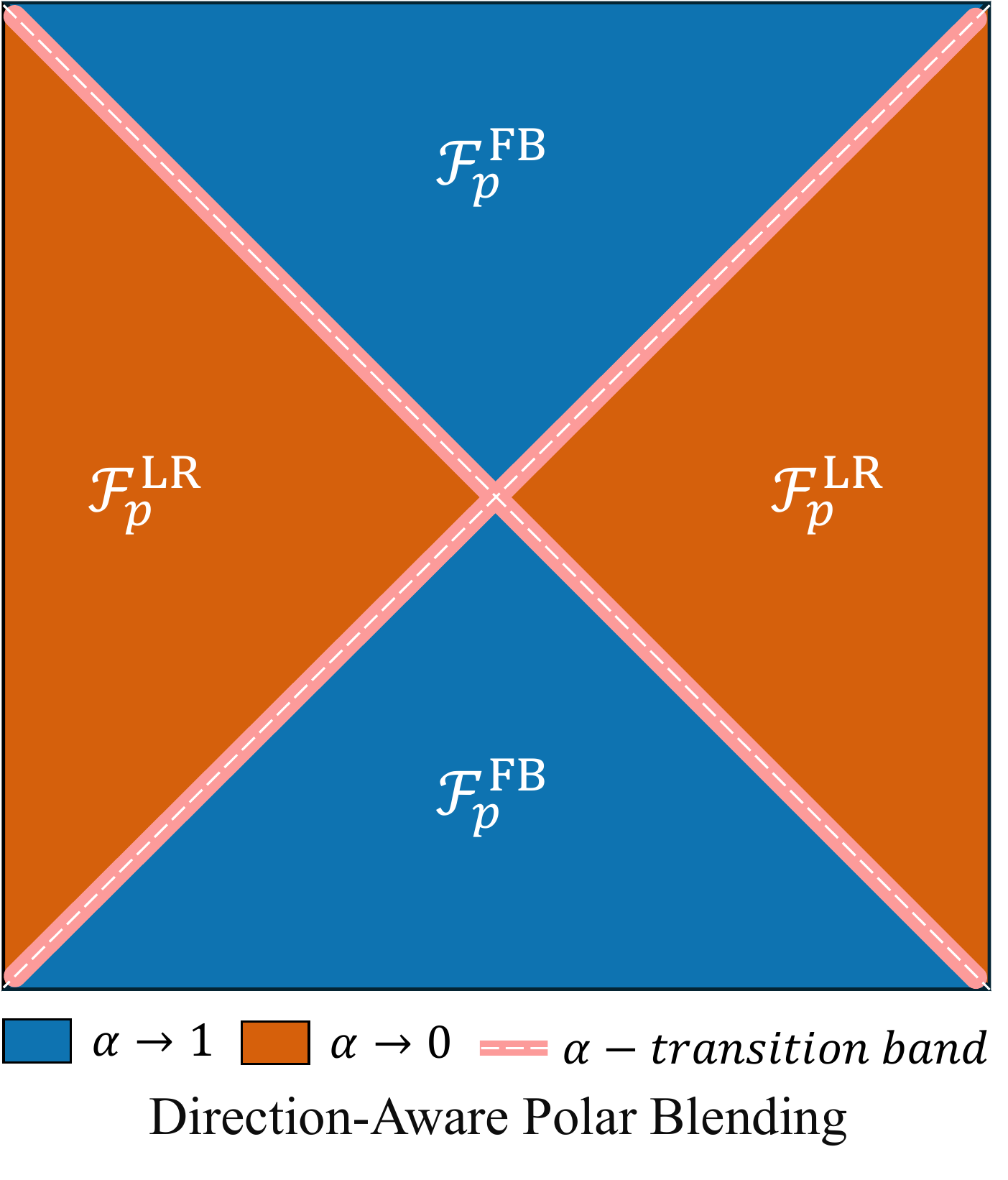}
\vspace{-0.15in}
\caption{\textbf{The direction-aware polar blending mask.}\label{fig:polar-blend-mask}}
\vspace{-0.25in}
\end{wrapfigure}
\paragraph{Representation Conversion.}
During training, panoramas are represented and supervised directly in CubeAtlas space.
To remain compatible with standard panorama rendering, storage, and evaluation pipelines, a generated CubeAtlas is deterministically converted into a standard cubemap and subsequently into ERP.
The equatorial band directly provides the four lateral faces, while the rotated half-regions in the upper and lower bands provide complementary observations for reconstructing the two polar faces.
All polar reconstruction and blending operations are performed in pixel space after decoding the CubeAtlas latents.

We first form two complementary full-face fusions.
For the upper pole, the front- and back-aligned half-regions $(\mathrm{U}_{\mathrm{F}},\mathrm{U}_{\mathrm{B}})$ are rotated back to their canonical orientations and joined to form a front--back (FB) estimate $\mathcal{F}_{\mathrm{U}}^{\mathrm{FB}}$.
Similarly, $(\mathrm{U}_{\mathrm{L}},\mathrm{U}_{\mathrm{R}})$ form a left--right (LR) estimate $\mathcal{F}_{\mathrm{U}}^{\mathrm{LR}}$.
Because each pair contains complementary halves of the same polar face, joining them along their shared boundary recovers a complete face.
We construct $\mathcal{F}_{\mathrm{D}}^{\mathrm{FB}}$ and $\mathcal{F}_{\mathrm{D}}^{\mathrm{LR}}$ for the lower pole in the same manner.
The resulting FB and LR fusions are geometrically aligned estimates of the same polar face, supported by complementary directional observations.
Although joint CubeAtlas generation already encourages them to be consistent, local differences may remain around complex structures near the polar center.
We therefore combine two estimates using the direction-aware blending rule as illustrated in \Cref{fig:polar-blend-mask}.
For each reconstructed polar face, let $(u,v)\in[-1,1]^2$ denote its normalized pixel coordinates.
For $p\in\{\mathrm{U},\mathrm{D}\}$, the rule favors $\mathcal{F}_{p}^{\mathrm{FB}}$ where $\lvert v\rvert>\lvert u\rvert$ and $\mathcal{F}_{p}^{\mathrm{LR}}$ where $\lvert u\rvert>\lvert v\rvert$, while smoothly blending them near the diagonals $\lvert u\rvert\approx\lvert v\rvert$.
With blending width $\tau$ and $[\cdot]_0^1=\mathrm{clip}(\cdot,0,1)$, we obtain:
\begin{equation}
\label{eq:polar}
\alpha(u,v)=
\left[
\frac{\lvert v\rvert-\lvert u\rvert+\tau}{2\tau}
\right]_0^1,
\qquad
\mathcal{F}_{p}
=
\alpha\,\mathcal{F}_{p}^{\mathrm{FB}}
+
(1-\alpha)\,\mathcal{F}_{p}^{\mathrm{LR}},
\quad p\in\{\mathrm{U},\mathrm{D}\}.
\end{equation}
This fusion preserves the directional consistency of all four polar observations and provides a smooth transition between the FB- and LR-dominant regions.
The reconstructed upper and lower faces, together with the four equatorial faces, form a standard cubemap, which is finally reprojected into the output ERP panorama.

\subsection{Model Design}
\label{sec:train}
\paragraph{Geometry-Free Panorama Generation.}
We build \methodname on a pre-trained image-editing DiT. The conditioning image is consumed directly:
we neither estimate nor provide camera intrinsics, camera pose, or an explicit projection of the input onto the sphere.
Instead, the model learns from data how to place the input content on the $360^{\circ}$ CubeAtlas layout and synthesize its surrounding context.

The conditioning image follows the standard dual-path design.
It is jointly encoded with the text by the frozen multimodal encoder~\citep{qwen25vl} for semantic context, and independently encoded by the frozen VAE into clean visual condition tokens.
These condition tokens are concatenated with noisy CubeAtlas target tokens, and the DiT operates on the joint
text, target, and condition sequence.
We fully fine-tune the DiT using the standard linear flow-matching objective, which predicts $\epsilon-x_0$ from $x_\sigma=(1-\sigma)x_0+\sigma\epsilon$, with $\sigma$ sampled from a logit-normal distribution.
We make no changes to the DiT architecture or its rotary position embeddings.
Consequently, \methodname preserves the strong perspective-image prior of the base model while adapting it to joint panorama generation in CubeAtlas space.

\paragraph{Layout-Aware Sparse Attention.}
Although the CubeAtlas canvas has a token count comparable to that of ERP, dense attention over all its regions is not always necessary to maintain panoramic consistency. 
A key advantage of CubeAtlas is that it makes geometrically adjacent faces locally connected while retaining a structured decomposition of the panorama.
We exploit this property with a layout-aware sparse attention mask over target-to-target interactions; text and condition tokens remain globally connected to every target token.

Specifically, we partition the target into upper, equatorial, and lower bands.
We retain full attention within each band to preserve horizontal consistency, but block direct interactions between the upper and lower polar bands.
For cross-latitude communication, each equatorial face attends only to the polar half-regions aligned with its axis.
For example, $\mathrm{L}$ attends to $\{\mathrm{U}_{\mathrm{L}},\mathrm{U}_{\mathrm{R}}, \mathrm{D}_{\mathrm{L}},\mathrm{D}_{\mathrm{R}}\}$, whose complementary halves reconstruct its neighboring upper and lower faces.
Conversely, a polar query attends to all equatorial keys, allowing the equatorial band to serve as the
cross-latitude bridge. 
Overall, the layout-aware mask retains $75\%$ of the target-to-target attention connectivity of dense attention, as illustrated in \Cref{fig:pipeline}, while text and image-conditioning tokens remain globally connected to all target tokens.
Blocking direct attention between some regions does not prevent information exchange: features can still propagate through the equatorial band across successive DiT layers. The mask therefore structures cross-region communication rather than separating the panorama into independently modeled regions. Our experiments further show that this structured sparsity supports cross-view consistency while improving efficiency.

\paragraph{Panoramic Super-Resolution.}
To obtain high-resolution panoramas without heavy training, we apply a frozen $4\times$ Pixel Diffusion Decoder (PiD)~\citep{lu2026pid} in perspective space. 
Specifically, the fused ERP is reprojected into six overlapping cubemap views with an enlarged FoV of $92.79^\circ$, which are independently enhanced by PiD and mapped back to the ERP domain.
This process upsamples the base panorama by $4\times$ along each spatial dimension, producing 10k ERP panoramas.
We linearly blend overlapping regions outside the FoV range of $90^\circ$ to suppress cross-view discontinuities.
Since generative super-resolution may introduce low-frequency shifts, we further apply a correction strategy~\citep{yue2025arbitrary} to anchor the global color and illumination to the base panorama while retaining the generated high-frequency details.

\section{Experiments}
\label{sec:exp}
\paragraph{Implementation Details.}
\label{sec:exp-setup}
We initialize \methodname from ~\citet{qwenedit}.
The VAE and Qwen2.5-VL condition encoder are kept frozen, while the denoising Transformer is fully fine-tuned for $8{,}000$ steps with a global batch size of $64$ and a learning rate of $10^{-5}$.
We apply FlexAttention~\citep{pytorch_flexattention} for the sparse attention implementation. 
Each cubemap face in CubeAtlas has a resolution of $640\times640$, corresponding to a fused ERP panorama of $1280\times2560$.
The super-resolution stage further upsamples the panorama by $4\times$ along each spatial dimension, producing a $5120\times10240$ panorama with a horizontal resolution of 10K.
Our evaluation benchmark consists of $100$ self-collected images, together with $100$, $100$, and $118$ panoramas from the RealSee3D~\citep{Li2025realsee3d_data}, Laval-Indoor~\citep{laval}, and SUN360~\citep{sun360} test sets, respectively.
We compare against CubeDiff~\citep{cubediff}, DreamCube~\citep{dreamcube}, HunyuanPano~\citep{hunyuanworld1}, and HY-Pano-2.0~\citep{hunyuanworld2} using their official inference configurations.
Unless otherwise specified, all comparisons are conducted on the base-resolution outputs without super-resolution refinement, ensuring a fair comparison across methods.
Additional datasets and evaluation details are provided in~\Cref{sec:impl}.

\subsection{Main Results}
\label{sec:main-results}
\paragraph{Quantitative Results.}
\Cref{tab:main_results} compares \methodname with representative panorama generation systems. 
The metrics cover distributional fidelity (FID, KID, and CLIP-FID), ERP and cubemap boundary continuity (Seam$_\text{ERP}$, Seam$_\text{Cube}$), equatorial and polar perceptual and geometric quality (Eq-* and Pole-*), and prompt adherence (TextAdh); full definitions are provided in \Cref{sec:metrics}.
\methodname achieves the lowest Seam$_\text{Cube}$ on all four benchmarks, reducing the cube-edge error by $63$--$78\%$ relative to CubeDiff.
These results support the effectiveness of generating a single canvas directly to eliminate cross-face discontinuities.
Note that at Seam$_{\text{ERP}}$ values around $0.01$, further numerical differences may not translate into perceptible differences in boundary seams. We therefore interpret this metric together with the qualitative comparisons.
\methodname also leads Eq-GI and Pole-GI on every benchmark and ranks first or second in Pole-IQ, demonstrating improved equatorial and polar geometry.
Meanwhile, it achieves the best CLIP-FID on all three public benchmarks, leads FID and KID on RealSee3D and FID on Laval-Indoor, and obtains the highest TextAdh.
These results confirm that \methodname improves spherical structure without sacrificing visual realism or textual consistency.

\begin{table*}[t]
\centering
\vspace{-0.15in}
\caption{{\textbf{Quantitative image-to-panorama comparison} on our self-collected set, RealSee3D, Laval-Indoor, and SUN360. 
FID, KID, and CLIP-FID are evaluated by $90^{\circ}$ perspective views split along $12$ icosahedral directions (eight valid directions for Laval-Indoor) and are not reported on the self-collected set without ground-truth ERPs. Seam$_\text{Cube}$ is reported only for methods with native cube-face outputs. 
}}
\label{tab:main_results}
\setlength{\tabcolsep}{3pt}
\resizebox{\textwidth}{!}{
\begin{tabular}{lccccccccccc}
\toprule
Method & FID~$\downarrow$ & KID~$\downarrow$ & CLIP-FID~$\downarrow$ & Seam$_\text{ERP}$~$\downarrow$ & Seam$_\text{Cube}$~$\downarrow$ & Eq-IQ~$\uparrow$ & Eq-Aes~$\uparrow$ & Eq-GI~$\uparrow$ & Pole-IQ~$\uparrow$ & Pole-GI~$\uparrow$ & TextAdh~$\uparrow$ \\
\midrule
\multicolumn{12}{l}{\textit{Self-Collected}} \\
\midrule
CubeDiff & - & - & - & \textbf{0.0126} & \underline{0.0361} & 3.685 & 3.565 & 3.465 & 3.780 & 3.725 & 4.100 \\
DreamCube & - & - & - & 0.0279 & 0.0445 & 2.730 & 2.605 & 2.607 & 2.870 & 2.935 & 3.490 \\
{HunyuanPano} & - & - & - & 0.0154 & - & 4.475 & 4.247 & 4.468 & 4.495 & 4.120 & 4.300 \\
{HY-Pano-2.0} & - & - & - & 0.0199 & - & \underline{4.580} & \underline{4.355} & \underline{4.480} & \textbf{4.705} & \underline{4.685} & \underline{4.470} \\
\textbf{Ours} & - & - & - & \underline{0.0142} & \textbf{0.0132} & \textbf{4.690} & \textbf{4.402} & \textbf{4.782} & \underline{4.675} & \textbf{4.730} & \textbf{4.640} \\
\midrule
\multicolumn{12}{l}{\textit{RealSee3D}} \\
\midrule
CubeDiff & 47.10 & 0.0085 & 5.60 & \underline{0.0083} & \underline{0.0229} & 3.217 & 2.803 & 3.375 & 3.700 & 3.770 & 3.060 \\
DreamCube & 61.66 & 0.0189 & 10.93 & 0.0189 & 0.0286 & 2.960 & 2.715 & 2.980 & 3.360 & 3.500 & 3.370 \\
{HunyuanPano} & 42.05 & 0.0068 & 4.61 & \textbf{0.0080} & - & 4.093 & 3.415 & 4.450 & 4.400 & 4.175 & 3.120 \\
{HY-Pano-2.0} & \underline{38.96} & \underline{0.0050} & \underline{3.00} & 0.0121 & - & \textbf{4.758} & \textbf{4.250} & \underline{4.787} & \underline{4.795} & \underline{4.735} & \underline{4.380} \\
\textbf{Ours} & \textbf{35.54} & \textbf{0.0036} & \textbf{2.50} & 0.0085 & \textbf{0.0064} & \underline{4.723} & \underline{4.062} & \textbf{4.867} & \textbf{4.820} & \textbf{4.795} & \textbf{4.620} \\
\midrule
\multicolumn{12}{l}{\textit{Laval-Indoor}} \\
\midrule
CubeDiff & 55.17 & 0.0082 & 7.75 & \underline{0.0064} & \underline{0.0181} & 3.235 & 2.665 & 3.385 & 3.815 & 3.900 & 3.450 \\
DreamCube & 92.06 & 0.0334 & 22.14 & 0.0245 & 0.0306 & 2.658 & 2.290 & 2.708 & 3.130 & 3.355 & 3.350 \\
{HunyuanPano} & 45.58 & 0.0049 & \underline{4.31} & \textbf{0.0051} & - & 3.895 & 3.277 & \underline{4.548} & 4.430 & 4.105 & 3.760 \\
{HY-Pano-2.0} & \underline{45.34} & \textbf{0.0034} & 6.08 & 0.0094 & - & \underline{4.228} & \textbf{3.635} & 4.482 & \textbf{4.740} & \underline{4.680} & \underline{4.480} \\
\textbf{Ours} & \textbf{43.37} & \underline{0.0037} & \textbf{4.03} & 0.0070 & \textbf{0.0039} & \textbf{4.237} & \underline{3.625} & \textbf{4.750} & \underline{4.665} & \textbf{4.710} & \textbf{4.670} \\
\midrule
\multicolumn{12}{l}{\textit{SUN360}} \\
\midrule
CubeDiff & 75.41 & 0.0318 & 12.30 & 0.0125 & 0.0315 & 2.557 & 2.284 & 2.716 & 3.123 & 3.089 & 3.051 \\
DreamCube & 66.28 & 0.0301 & 14.05 & 0.0192 & \underline{0.0281} & 2.617 & 2.445 & 2.737 & 3.267 & 3.483 & 3.508 \\
{HunyuanPano} & \underline{37.28} & \underline{0.0103} & 7.11 & \textbf{0.0069} & - & 3.710 & \underline{3.364} & 4.059 & 4.072 & 3.877 & 3.441 \\
{HY-Pano-2.0} & \textbf{36.71} & \textbf{0.0076} & \underline{5.75} & 0.0106 & - & \underline{4.017} & \textbf{3.843} & \underline{4.222} & \underline{4.496} & \underline{4.369} & \underline{4.424} \\
\textbf{Ours} & 38.70 & 0.0105 & \textbf{5.62} & \underline{0.0101} & \textbf{0.0068} & \textbf{4.184} & \textbf{3.843} & \textbf{4.379} & \textbf{4.530} & \textbf{4.470} & \textbf{4.661} \\
\bottomrule
\end{tabular}}
\vspace{-0.15in}
\end{table*}

\begin{figure}
\begin{center}
\includegraphics[width=1.0\linewidth]{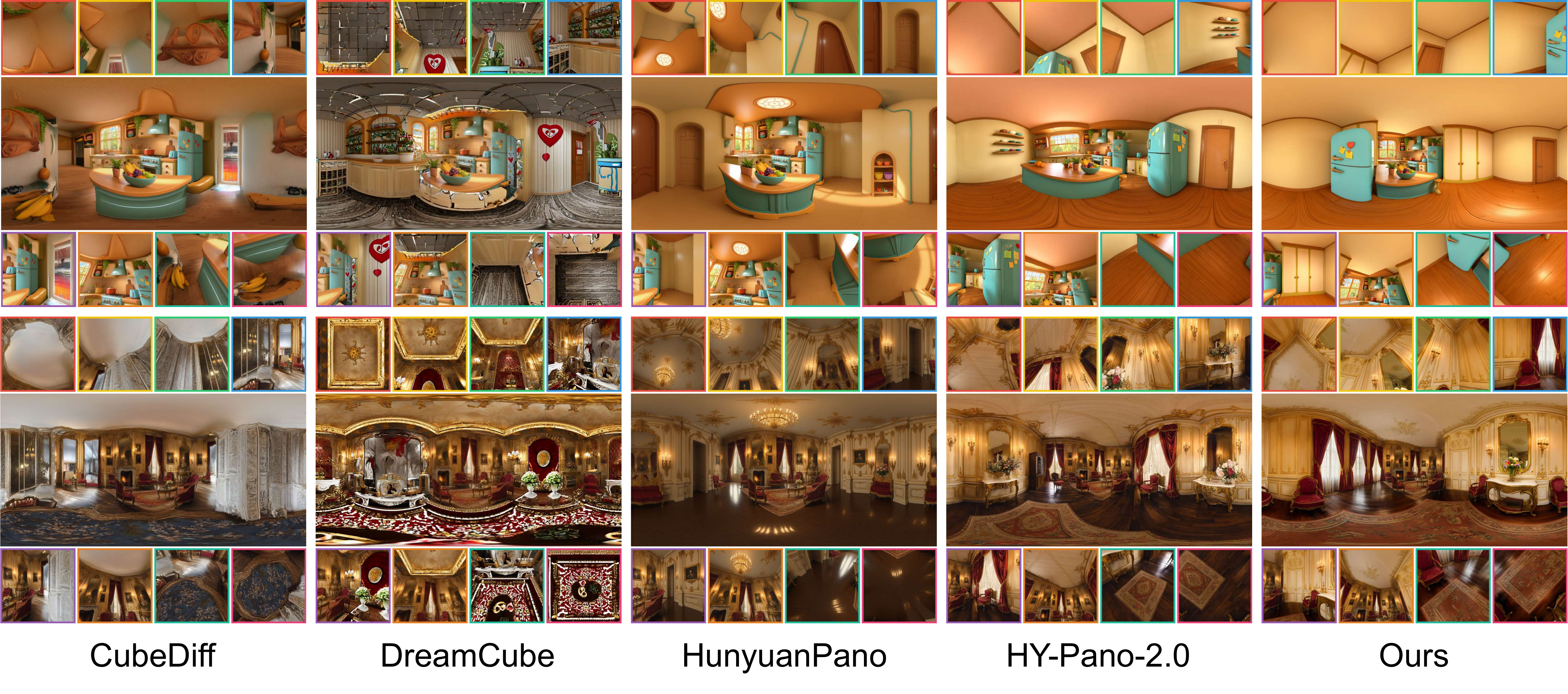}
\end{center}
\vspace{-0.15in}
\caption{{\textbf{Qualitative comparison with perspective cubemaps.}
Our results enjoy superior cross-face structures and pole geometry compared to ERP-native and cubemap-based baselines.}\label{fig:qualitative}}
\vspace{-0.15in}
\end{figure}

\paragraph{Qualitative Results.}
\Cref{fig:qualitative} compares \methodname with CubeDiff, DreamCube, HunyuanPano, and HY-Pano-2.0 using perspective cubemap views.
Across diverse indoor and outdoor scenes, \methodname maintains coherent geometry and lighting across both the ERP wrap boundary and adjacent cubemap views, while conventional cubemap-based methods often exhibit visible face seams. 
In the upper and lower polar regions, it also substantially reduces the radial stretching, pinching, and structural collapse typical of direct ERP generation.
These visual improvements align with the seam-continuity and regional-geometry results in \Cref{tab:main_results}.

\begin{table*}
\centering
\vspace{-0.15in}
\caption{\textbf{Ablation on panorama representation and efficiency-oriented designs.} ``Tokens'' denotes the target-token sequence length, and ``Time'' denotes denoising time under the same settings. Both are reported as relative multipliers normalized to the ERP baseline ($1.00\times$). ``Sparse Attn.'' indicates the layout-aware sparse attention used by \methodname.}
\label{tab:ablation}
\setlength{\tabcolsep}{4pt}
\resizebox{\textwidth}{!}{
\begin{tabular}{lcccccccccc}
\toprule
& \multicolumn{2}{c}{Computational Profile} & \multicolumn{8}{c}{Quality and Continuity} \\
\cmidrule(lr){2-3} \cmidrule(lr){4-11}
Variant & Tokens & Time & $\text{Seam}_{\text{ERP}}\downarrow$ & $\text{Seam}_{\text{Cube}}\downarrow$ & Eq-IQ$\uparrow$ & Eq-Aes$\uparrow$ & Eq-GI$\uparrow$ & Pole-IQ$\uparrow$ & Pole-GI$\uparrow$ & TextAdh$\uparrow$ \\
\midrule
ERP & $1.00\times$ & $1.00\times$ & 0.0209 & -- & 4.607 & \underline{4.370} & 4.575 & \underline{4.690} & 4.660 & 4.550 \\
Cubemap & $0.75\times$ & $0.76\times$ & 0.0147 & 0.0260 & 4.428 & 4.270 & 4.332 & 4.490 & 4.455 & \underline{4.560} \\
\midrule
CubeAtlas & $1.00\times$ & $1.00\times$ & \textbf{0.0103} & \underline{0.0143} & \textbf{4.697} & 4.353 & \underline{4.690} & \textbf{4.705} & \underline{4.705} & \textbf{4.640} \\
\quad + Sparse Attn. & $1.00\times$ & $0.84\times$ & \underline{0.0142} & \textbf{0.0132} & \underline{4.690} & \textbf{4.402} & \textbf{4.782} & 4.675 & \textbf{4.730} & \textbf{4.640} \\
\bottomrule
\end{tabular}}
\vspace{-0.15in}
\end{table*}

\subsection{Ablation}
\label{sec:ablation}

\begin{figure*}
\centering
\includegraphics[width=1.0\textwidth]{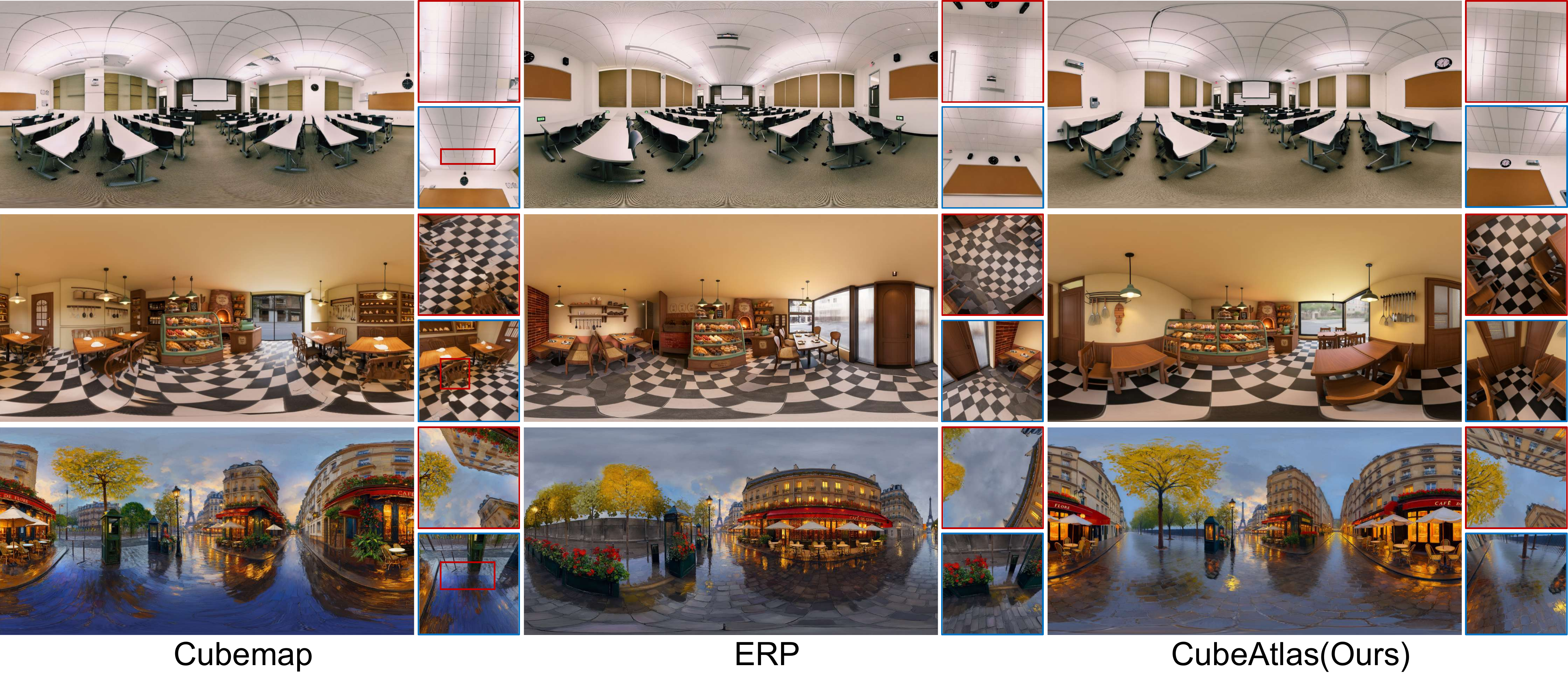}
\vspace{-0.15in}
\caption{{\textbf{Visual comparison of panorama representations.} CubeAtlas
reduces cubemap face discontinuities while preserving polar structures more
faithfully than direct ERP generation.}\label{fig:ablation-qualitative}}
\vspace{-0.15in}
\end{figure*}

\noindent\textbf{Panorama Representation.}
We first compare our CubeAtlas with ERP and cubemap generation, while keeping the backbone and other training settings fixed.
Both the ERP baseline and CubeAtlas use the same horizontal padding scheme, while all results are obtained without super-resolution or additional restoration for fair comparisons.
As shown in \Cref{tab:ablation}, cubemap avoids ERP's polar distortion but suffers from serious cube-boundary inconsistency. 
CubeAtlas substantially reduces this error and achieves the strongest Eq-IQ and Pole-IQ, together with competitive distortion scores.
The visual comparison in \Cref{fig:ablation-qualitative} further shows that CubeAtlas reduces both cross-face discontinuities in the cubemap and polar deformation in ERP.

\noindent\textbf{Layout-Aware Sparse Attention.}
We compare dense and sparse attention under the same CubeAtlas representation.
As shown in \Cref{tab:ablation}, sparse attention preserves comparable overall generation quality: it reduces Seam$_\text{Cube}$ and improves Eq-Aes, Eq-GI, and Pole-GI, while incurring only minor degradations in Eq-IQ and Pole-IQ and a modest increase in Seam$_\text{ERP}$. By reducing unnecessary target-to-target interactions, the sparse mask lowers the attention cost of CubeAtlas without changing its token count. 
Across representations, cubemap remains the most efficient due to its smaller token budget, whereas CubeAtlas is more efficient than ERP at the same token count. Importantly, CubeAtlas delivers consistently better generation quality than both representations, as shown quantitatively in \Cref{tab:ablation} and qualitatively in
\Cref{fig:ablation-qualitative}.

\begin{figure*}
\centering
\includegraphics[width=1.0\textwidth]{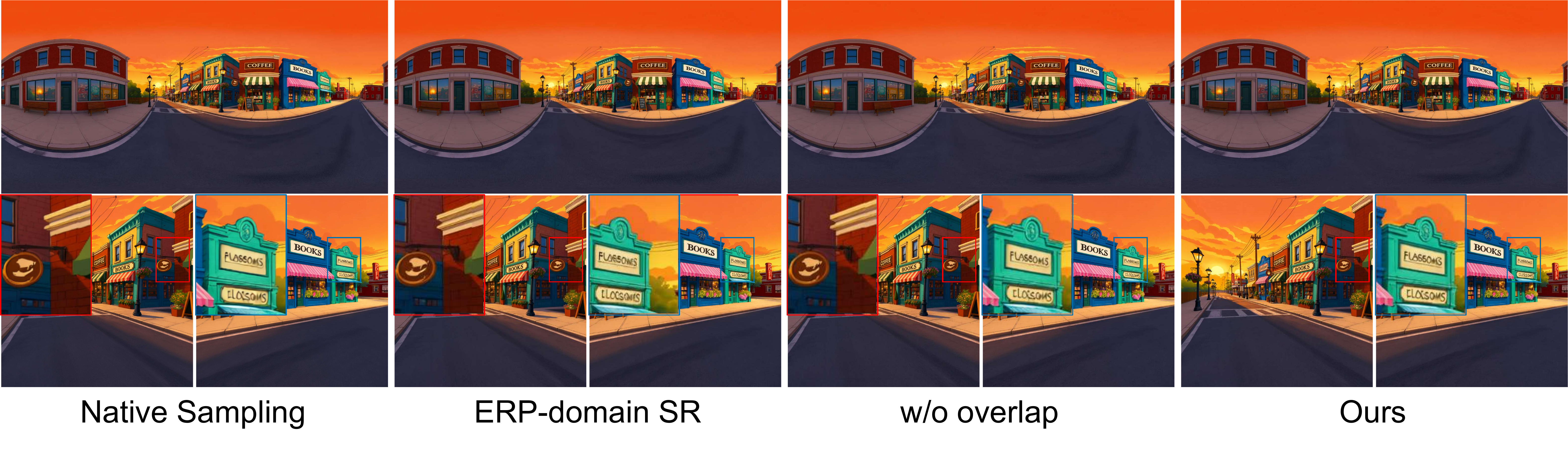}
\vspace{-0.15in}
\caption{\textbf{Qualitative comparison of panoramic super-resolution.}
Our method recovers sharper details while maintaining cross-face consistency.\label{fig:qualitative-sr}}
\vspace{-0.15in}
\end{figure*}

\noindent\textbf{Panoramic Super-Resolution.}
\label{sec:ablation-sr}
\Cref{fig:qualitative-sr} compares native interpolation-based sampling, ERP-based super-resolution,
perspective super-resolution without overlap fusion, and our full pipeline.
The interpolation-based sampling lacks fine detail, while ERP-based processing produces uneven
enhancement because of spherical distortion. Processing cubemap views without
overlap fusion sharpens local details but introduces inconsistencies across face
boundaries. Our full pipeline recovers sharper structures and textures while
maintaining a coherent appearance across the panorama.

\section{Conclusion}
\label{sec:conclusion}
We presented \methodname, a representation-driven framework for text-guided
image-to-panorama generation. Its CubeAtlas representation jointly generates a
compact grid of equatorial faces and rotated half-polar regions, combining
cubemap perspective geometry with global continuity while retaining an ERP-like
token count. Layout-aware sparse attention further removes unnecessary
cross-region interactions, and geometry-free conditioning transfers pretrained
image priors without camera calibration or architectural modification. A
training-free perspective-space super-resolution stage produces the final
high-resolution ERP. Across diverse benchmarks, \methodname improves seam
continuity and spherical geometry while preserving visual fidelity and text
adherence. These results highlight panorama representation as a simple and
effective route for adapting generative image models to omnidirectional synthesis.

\subsection*{AI use statement}

In this work, we used generative AI tools to assist with manuscript writing and language editing. We also used generative AI tools to generate textual captions for training and evaluation, and to perform automated evaluation. We have reviewed all AI-assisted work. We take responsibility for the final content of this work, including text, claims or artifacts produced with the aid of generative AI.

\bibliography{iclr2027_conference}
\bibliographystyle{iclr2027_conference}

\appendix
\raggedbottom
\section*{Appendix}
\section{Limitations}
Our current model is trained and evaluated at a fixed base resolution and does not explicitly support arbitrary-resolution generation.
In practice, however, the perspective-space super-resolution pipeline produces panoramas up to $5120\times10240$ resolution, which is sufficient for most downstream applications and makes output resolution unlikely to be a practical bottleneck.
As a generative model, \methodname also depends on the quality of its textual conditioning.
We observe that accurate and detailed captions provide substantially stronger semantic guidance, particularly for regions that are not visible in the input perspective image, whereas vague or incomplete descriptions may lead to less coherent or less controllable generations.
More fundamentally, generating a complete $360^\circ$ environment from a single perspective observation is inherently under-constrained.
Content outside the observed FoV must be synthesized rather than faithfully recovered, and multiple plausible panoramas may correspond to the same input image. 
Finally, our current study focuses on static panoramas. Extending CubeAtlas to temporally consistent panoramic video and interactive scene generation remains an interesting direction for future work.

\section{Supplementary Implementation Details}
\label{sec:impl}

\paragraph{Training Datasets.}
Our training corpus combines publicly available, internally collected, and synthetic panoramas. The public portion includes PolyHaven~\citep{polyhaven}, Humus~\citep{humus}, and PanoInfinigen~\citep{bozic2026pager}, together with additional street-view panoramas collected from Mapillary~\citep{mapillary}.
After quality filtering, these public sources provide approximately 50K panoramas.
We further include 20K internally collected panoramas covering street scenes and indoor environments, and render 5K synthetic data from Unreal Engine assets following the pipeline in~\cite {lu2025genex}. 
Together, these sources form a training corpus of roughly 80K panoramas.
We sample real and synthetic panoramas during training at an approximate ratio of $6{:}4$.

\paragraph{Evaluation Datasets.}
Our evaluation suite consists of three public panorama benchmarks and one self-collected image set. Specifically, we evaluate on $100$ panoramas from RealSee3D~\citep{Li2025realsee3d_data}, $100$ panoramas from the Laval-Indoor test set~\citep{laval}, and $118$ panoramas from the SUN360 test set~\citep{sun360}. For each panorama, we render a $90^\circ$-FoV perspective view centered at the canonical viewing direction as the conditioning image.
Gemini-3.5-Flash generates the text prompt from the complete ground-truth panorama, providing a description of both the observed and unobserved regions. This setting therefore evaluates panorama generation under relatively complete scene-level textual guidance.

Our self-collected set provides a more challenging evaluation setting. It contains $100$ perspective images spanning diverse visual styles, with $50$ indoor and $50$ outdoor scenes. Unlike the public benchmarks, no ground-truth panorama is available, and the text prompt is generated solely from the conditioning image. The model must therefore infer plausible content for the unobserved surroundings rather than relying on textual descriptions of the
complete scene. Since ground-truth ERP images are unavailable for this set, we report only reference-free metrics.

Importantly, none of the evaluation images are used during training. The same separation is maintained for all ablation models, ensuring that both the main model and all ablated variants are evaluated on entirely unseen data.

\paragraph{Formulation Details of CubeAtlas.}
Training targets for \methodname are constructed online in CubeAtlas space. Before VAE encoding, a $32$-pixel-wide strip copied from the left edge is appended to the right boundary as auxiliary horizontal context. This padding facilitates wrap-around seam handling but does not belong to the core CubeAtlas representation. During output assembly, the decoded padding context is used for boundary blending and then removed. The ERP ablation uses the same context-padding scheme for a fair comparison. Sparse attention is applied only to target-to-target interactions, while text and image-conditioning tokens remain globally connected.

\paragraph{Condition Sampling and Captioning.}
For each training panorama, we render a perspective conditioning view by uniformly sampling the FoV from $[50^\circ, 90^\circ]$ and the pitch angle from $[-75^\circ, 75^\circ]$, while fixing the yaw and roll angles to $0^\circ$. To improve robustness to diverse input formats, we render the conditioning image at a nominal resolution of 1024 pixels, with its aspect ratio sampled from $\{3\!:\!4, 1\!:\!1, 4\!:\!3, 16\!:\!9, 9\!:\!16\}$. For text conditioning, we use Gemini-3.5-Flash to describe the complete panorama rather than the sampled perspective view. For each panorama, we generate both a concise and a detailed caption and randomly use one of them during training, exposing the model to scene-level descriptions at different levels of granularity.

\paragraph{Optimization Details.}
We train with AdamW using a global batch size of $64$, a learning rate of $10^{-5}$, betas $(0.9,0.999)$, weight decay $0.01$, and gradient clipping at $1.0$. The learning rate is linearly warmed up for $100$ optimization steps and
then kept constant. We train our final model for $8{,}000$ steps using exponential moving average (EMA). Training uses BF16 forward computation, FP32 master weights and gradient reduction, and activation checkpointing. 
All experiments are based on 64 H20 NVIDIA GPUs.

\paragraph{Super-Resolution Details.}
We use frozen Flux-based PiD weights~\citep{lu2026pid} without panorama-specific fine-tuning. 
To enable seamless fusion after super-resolution, the base ERP is reprojected into six overlapping
perspective cubemap views by slightly enlarging the FoV to $92.79^\circ$.
Each view has a
spatial resolution of $672\times672$, where the additional context beyond the canonical $90^\circ$
cubemap face forms the overlap regions with neighboring views. PiD independently upsamples each
view by $4\times$ to $2688\times2688$, so the overlap regions are enhanced together with the valid
face content. The super-resolved views are then reprojected back to the ERP domain, where overlapping
regions are linearly blended to suppress cross-view discontinuities. After removing the corresponding 128px boundary context, each effective cube face has a resolution of $2560\times2560$,
yielding a final ERP panorama of $5120\times10240$.

\section{User Study}
\label{sec:user-study}
\begin{figure}[t]
\centering
\includegraphics[width=\linewidth]{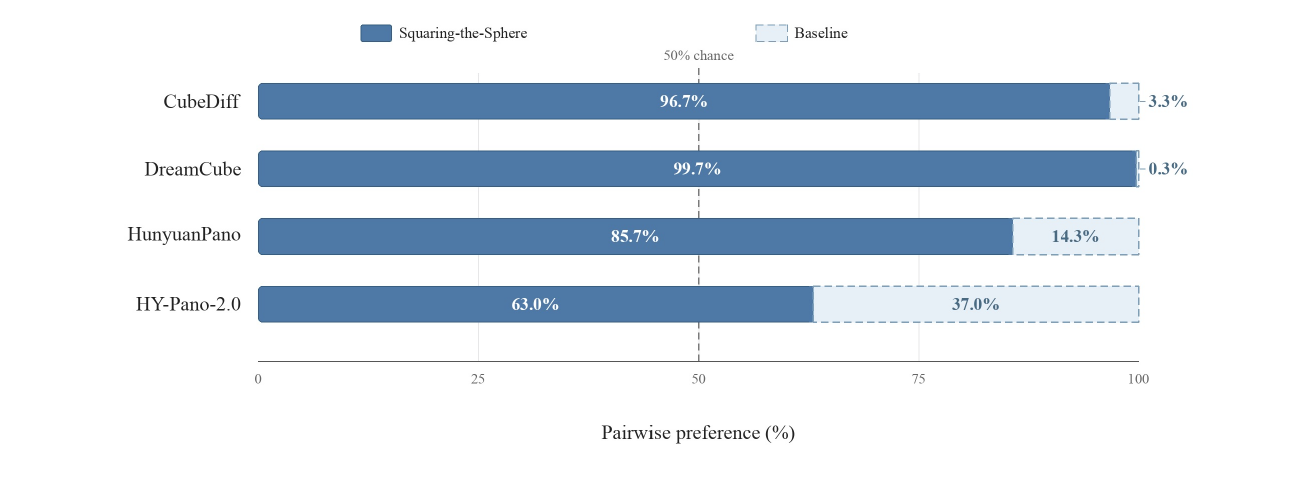}
\vspace{-0.15in}
\caption{\textbf{User study.} 
Each stacked bar summarizes $700$ valid judgments comparing \methodname (solid blue) with one competitor (light fill with a dashed border). The vertical dashed line denotes the $50\%$ equal-preference reference.\label{fig:user-study}}
\vspace{-0.15in}
\end{figure}

We conduct a two-alternative forced-choice user study with $28$ volunteers.
Participants are divided into four groups of seven, with each group comparing \methodname against one competing method. Each participant evaluates all $100$ scenes assigned to their group, resulting in $700$ valid comparisons per competing method and $2{,}800$ comparisons in total. Each trial requires a binary preference with no tie option.
For all methods, we use the base-resolution outputs without super-resolution. Compared methods are produced using their official implementations and inference settings. In each trial, participants are shown the conditioning image and text prompt together with two anonymized panorama results presented in randomized left--right order. Both panoramas can be freely explored using an interactive
$360^\circ$ viewer. Participants are asked to select the better result by jointly considering conditioning fidelity, visual quality, boundary continuity, and polar geometry.
As shown in \Cref{fig:user-study}, \methodname is preferred over CubeDiff, DreamCube, HunyuanPano, and HY-Pano-2.0 in $96.7\%$, $99.7\%$, $85.7\%$, and $63.0\%$ of comparisons, respectively.

\section{Evaluation Metrics}
\label{sec:metrics}
We evaluate image-to-panorama generation along four complementary dimensions: distributional fidelity, boundary continuity, regional perceptual and geometric quality, and text adherence. 

\paragraph{Distributional Fidelity.}
\label{sec:metrics_views}
We evaluate distributional fidelity using FID~\citep{heusel2017fid}, KID~\citep{binkowski2018kid}, and CLIP-FID~\citep{kynkaanniemi2023cleanfid}.
Because their feature encoders are trained on perspective images, we compute these
metrics on tangent perspective views rather than directly on ERP panoramas.
For each benchmark, both generated and ground-truth panoramas are resized to the
native resolution of the ground-truth set and rendered into $90^\circ$ square views
centered at the $12$ vertices of a regular icosahedron. All rendered views are pooled
at the dataset level for metric computation. FID and KID use Inception-V3 features,
while CLIP-FID uses CLIP ViT-B/32 features; all metrics are computed with clean-fid~\cite{cleanfid}, with lower values indicating better performance.
KID is reported without additional scaling.

RealSee3D and SUN360 use all $12$ viewing directions. For Laval-Indoor, views
centered below $-23^\circ$ are excluded from both generated and ground-truth sets
due to invalid lower regions in the ground-truth panoramas, leaving eight valid
directions. Distributional metrics are not reported on our self-collected set because
ground-truth panoramas are unavailable.

\paragraph{Boundary Continuity.}
\label{sec:metrics_seam}
We use two reference-free diagnostics for representation-specific seams, both
computed as RGB mean absolute error in normalized $[0,1]$ color space. For panorama
$I_i\in[0,1]^{H\times W\times3}$, the ERP wrap-around error is
\begin{align}
s_{\mathrm{ERP}}(I_i)
&=\frac{1}{3H}\sum_{h,c}\left|I_i(h,0,c)-I_i(h,W-1,c)\right|,\\
\mathrm{Seam}_{\mathrm{ERP}}
&=\frac{1}{N}\sum_{i=1}^{N}s_{\mathrm{ERP}}(I_i).
\end{align}
For Seam$_\mathrm{Cube}$, the six native cubemap faces are resized to
$512\times512$, aligned according to cubemap topology, and evaluated over all
$12$ shared edge pairs:
\begin{equation}
\mathrm{Seam}_{\mathrm{Cube}}=\frac{1}{12N}\sum_{i=1}^{N}
\sum_{e=1}^{12}\operatorname{MAE}\!\left(E_{i,e}^{(a)},E_{i,e}^{(b)}\right).
\end{equation}
Lower values indicate better boundary continuity. Seam$_\mathrm{Cube}$ is computed
from native faces and is therefore undefined for methods that output only ERP
images. Because these pixel-level diagnostics are also affected by local texture
variation, we interpret them together with the qualitative comparisons.

\paragraph{VLM-based Regional Quality and Text Adherence.}
\label{sec:metrics_vlm}
We use Gemini-3.5-Flash to evaluate regional visual quality and text adherence on a shared five-point scale: Excellent (5), Good (4), Fair (3), Poor (2), and Awful (1). Each generated ERP is resized to $2048\times1024$ before evaluation.
For the equatorial region, we render four $512\times512$ perspective views with a $90^\circ$ FoV at pitch $0^\circ$ and yaw
$\{0^\circ,90^\circ,180^\circ,270^\circ\}$, and concatenate them into a strip labeled Front, Right, Back, and Left. For the polar regions, we render two $512\times512$, $90^\circ$ perspective views centered at pitch $+90^\circ$ and $-90^\circ$, labeled Zenith and Nadir, and place them above the full ERP, which is provided only as contextual reference. The regional prompts ask the judge to evaluate each view separately and then provide a score. For each view and criterion, they require a brief rationale followed by an integer score.

Eq-IQ and Pole-IQ correspond to the ``Image Quality'' criterion, which asks for the overall quality of each equatorial view or polar close-up. Eq-Aes corresponds to ``Aesthetic'', which asks how visually pleasing each equatorial view is. Eq-GI and Pole-GI correspond to the ``Distortion'' criterion in the respective regional prompts and are reported as \emph{geometric integrity} for clarity, without changing the numerical scores. Higher scores indicate fewer distortion-related visual artifacts.
For Eq-GI, the prompt asks whether the view resembles a normal camera photo, with wavy or rippling lines, melted or broken objects, and shimmering moire or banding on regular textures lowering the score. For Pole-GI, the prompt asks whether the pole converges naturally and identifies swirl or vortex patterns, central black holes, radial cracks or seams, warped geometry, smearing, and nadir camera or tripod patches as defects. Thus, these criteria cover the specified geometric and structural artifacts as well as texture artifacts or polar patches, rather than a purely geometric reconstruction error. For each regional criterion, we first average scores over the corresponding views within each panorama, then over panoramas.

TextAdh uses the same judge and five-point scale, but evaluates the complete ERP together with its input caption. The prompt asks the judge to assess only whether the visible content matches the description, considering objects, attributes, counts, and spatial relations. Its anchors range from all described elements being present and correct (5), through minor omissions or small mismatches (4), several missing or incorrect elements (3), and most of the description being unsatisfied (2), to a scene nearly unrelated to the prompt (1). Scores are averaged over samples with non-empty prompts. The scoring prompts are provided below.

\paragraph{VLM Scoring Instructions.}
\label{sec:vlm-prompts}
We provide the task-specific scoring prompts for equatorial quality, polar quality, and text adherence below. Equatorial evaluation covers overall image quality, visual appeal, and distortion-related artifacts in perspective views, while polar evaluation covers close-up quality and the specified polar defects. Text adherence is assessed with a separate prompt focused on semantic matching.

\begin{figure}
\centering
\includegraphics[width=1.0\linewidth]{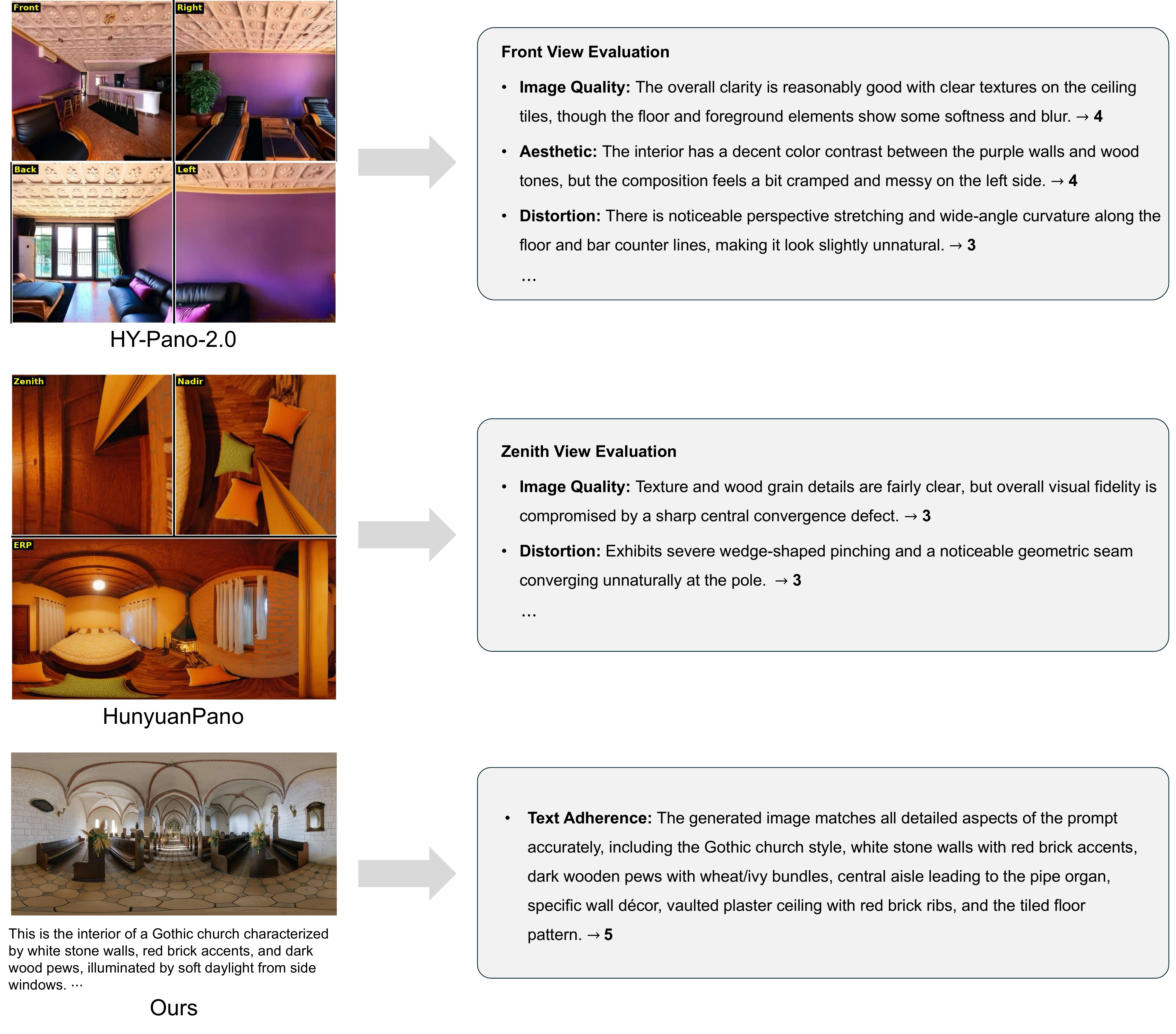}
\vspace{-0.15in}
\caption{\textbf{Examples of VLM-based evaluation.} Regional examples show rationales and scores for the Front and Zenith views. ``Distortion'' denotes geometric integrity (GI), with higher scores indicating fewer artifacts. Equatorial views are rearranged for display, and the input caption is excerpted.\label{fig:vlm-examples}}
\vspace{-0.15in}
\end{figure}

The examples in \Cref{fig:vlm-examples} illustrate how the VLM grounds its ratings in visible image content. From top to bottom, the figure presents equatorial quality, polar quality, and text-adherence evaluation, with input on the left and VLM rationales and scores on the right. Regional outputs are shown for selected views, while TextAdh evaluates the full panorama.

\begin{vlmprompt}{A. Equatorial Evaluation}
You are an evaluator assessing a panorama generation model.
The input is one stitched strip of four rectilinear eye-level perspective views labeled Front, Right, Back, Left, cropped from one generated 360-degree panorama. Evaluate EACH of the four views separately, and then provide a score based on the following three criteria:\newline
1. \textbf{Image Quality}: overall quality of the view.\newline
2. \textbf{Aesthetic}: how visually pleasing the view is.\newline
3. \textbf{Distortion}: does it look like a photo from a normal camera? Wavy/rippling lines,
melted or broken objects, or shimmering moire/banding on regular textures lower it.
\par\smallskip
Each criterion uses a five-point scale: Excellent (5), Good (4), Fair (3), Poor (2), Awful (1). For every view and criterion, give a brief reason first, then the integer score.
\par\smallskip
Forward your response as follows:\newline
- \textbf{Image Quality}: (Brief reason) $\rightarrow$ Score\newline
- \textbf{Aesthetic}: (Brief reason) $\rightarrow$ Score\newline
- \textbf{Distortion}: (Brief reason) $\rightarrow$ Score
\par\smallskip
\texttt{\textless IMAGE\textgreater}
\end{vlmprompt}

\begin{vlmprompt}{B. Polar Evaluation}
You are an evaluator assessing a panorama generation model.
The input shows two rectilinear 90-degree polar close-ups labeled Zenith (up) and Nadir (down) on top; the full equirectangular panorama (ERP) is below for context only. Evaluate EACH polar view separately, and then provide a score based on the following two criteria:\newline
1. \textbf{Image Quality}: overall quality of the close-up.\newline
2. \textbf{Distortion}: does the pole converge naturally, or is there swirl/vortex, a central black
hole, radial cracks/seams, warped geometry, smearing, or a nadir camera/tripod patch?
\par\smallskip
Each criterion uses a five-point scale: Excellent (5), Good (4), Fair (3), Poor (2), Awful (1). For every view and criterion, give a brief reason first, then the integer score.
\par\smallskip
Forward your response as follows:\newline
- \textbf{Image Quality}: (Brief reason) $\rightarrow$ Score\newline
- \textbf{Distortion}: (Brief reason) $\rightarrow$ Score
\par\smallskip
\texttt{\textless IMAGE\textgreater}
\end{vlmprompt}

\begin{vlmprompt}{C. Text Adherence}
You are evaluating how well a generated 360-degree panorama matches its text prompt.
The image is one equirectangular panorama (ERP) showing the full 360 scene at once; Judge ONLY whether the visible content matches the description below.
\par\smallskip
Prompt: "\texttt{\textless TEXT\_PROMPT\textgreater}"
\par\smallskip
Rate text adherence on a five-point scale --- Excellent (5): all described objects, attributes,
counts, and spatial relations are present and correct; Good (4): minor omissions or small
mismatches; Fair (3): several described elements missing or wrong; Poor (2): most of the
description is not satisfied; Awful (1): the scene is nearly unrelated to the prompt.
\par\smallskip
Forward your response as follows:\newline
- \textbf{Text Adherence}: (Brief reason) $\rightarrow$ Score
\par\smallskip
\texttt{\textless IMAGE\textgreater}
\end{vlmprompt}

\begin{table}
\centering
\caption{\textbf{Agreement between VLM scores and human judgments.}
Spearman rank correlations between VLM scores and human MOS on 30 generated panoramas, each evaluated by three raters. Higher values indicate stronger rank agreement with human judgments.}
\label{tab:vlm-human}
\small
\setlength{\tabcolsep}{8pt}
\begin{tabular}{lcccccc}
\toprule
Metric & Eq-IQ & Eq-Aes & Pole-IQ & Eq-GI & Pole-GI & TextAdh \\
\midrule
Spearman $\rho$ & 0.9358 & 0.8820 & 0.8757 & 0.9281 & 0.7698 & 0.8804 \\
\bottomrule
\end{tabular}
\end{table}

\paragraph{Agreement with Human Judgments.}
\label{sec:vlm-human}
To validate the VLM-based evaluation, we sample 10 test cases and select outputs from three randomly chosen methods per case, yielding 30 panoramas that collectively cover CubeDiff, DreamCube, HunyuanPano, HY-Pano-2.0, and our method. Three independent raters, blinded to method identities and VLM scores, evaluate each panorama on Eq-IQ, Eq-Aes, Pole-IQ, Eq-GI, Pole-GI, and TextAdh using a five-point scale. Regional panels are rated as a whole. We average the three human ratings to obtain a mean opinion score (MOS) for each panorama and criterion. For each criterion, we compute Spearman's rank correlation across the 30 panoramas between MOS and the corresponding panorama-level VLM scores.
As shown in \Cref{tab:vlm-human}, VLM scores correlate strongly with human judgments across all six criteria, with $\rho$ ranging from $0.7698$ to $0.9358$.

\section{Additional Results}

\subsection{Full-Polar vs Half-Polar Packing}
\label{sec:polar-packing-comparison}
\Cref{tab:polar-packing-comparison} compares a full-polar variant of \methodname with the half-polar design adopted in our final model. The full-polar variant retains a complete orientation-aligned copy of each polar face above and below every equatorial face, resulting in a fully occupied grid. Compared with the half-polar design, this increases the token count by $50\%$ and incurs approximately
$37\%$ higher computational cost (even enhanced by the sparse attention). All quality metrics and efficiency measurements in \Cref{tab:polar-packing-comparison} are evaluated on the self-collected benchmark. 

Despite its substantially higher computational cost, the full-polar variant does not yield a clear improvement over the half-polar design across the evaluated metrics. This suggests that retaining only the equator-adjacent halves already provides sufficient polar context for high-quality panorama generation, while avoiding the redundancy introduced by complete polar copies. The half-polar
packing therefore offers a superior quality-and-efficiency trade-off and is adopted in our final design.

\begin{table*}
\centering
\caption{\textbf{Full-polar vs half-polar packing.}
Padding tokens are omitted from the token length counting for simplification.
``+ Sparse Attn.'' denotes layout-aware sparse attention.}
\label{tab:polar-packing-comparison}
\setlength{\tabcolsep}{3pt}
\resizebox{\textwidth}{!}{
\begin{tabular}{lcccccccccc}
\toprule
& \multicolumn{2}{c}{Computational Profile} & \multicolumn{8}{c}{Quality and Continuity} \\
\cmidrule(lr){2-3} \cmidrule(lr){4-11}
Variant & Tokens & Time & Seam$_{\mathrm{ERP}}\downarrow$ & Seam$_{\mathrm{Cube}}\downarrow$ & Eq-IQ$\uparrow$ & Eq-Aes$\uparrow$ & Eq-GI$\uparrow$ & Pole-IQ$\uparrow$ & Pole-GI$\uparrow$ & TextAdh$\uparrow$ \\
\midrule
Half-polar & $1.00\times$ & $1.00\times$ & 0.0103 & 0.0143 & 4.697 & 4.353 & 4.690 & 4.705 & 4.705 & 4.640 \\
\quad + Sparse Attn. & $1.00\times$ & $0.84\times$ & 0.0142 & 0.0132 & 4.690 & 4.402 & 4.782 & 4.675 & 4.730 & 4.640 \\
\midrule
Full-polar & $1.50\times$ & $1.63\times$ & 0.0106 & 0.0134 & 4.610 & 4.410 & 4.670 & 4.660 & 4.710 & 4.580 \\
\quad + Sparse Attn. & $1.50\times$ & $1.15\times$ & 0.0103 & 0.0136 & 4.620 & 4.410 & 4.720 & 4.660 & 4.680 & 4.620 \\
\bottomrule
\end{tabular}}
\end{table*}

\subsection{Geometry-Free Generation from Arbitrary Views}

\begin{figure*}[p]
\centering
\includegraphics[width=0.95\textwidth]{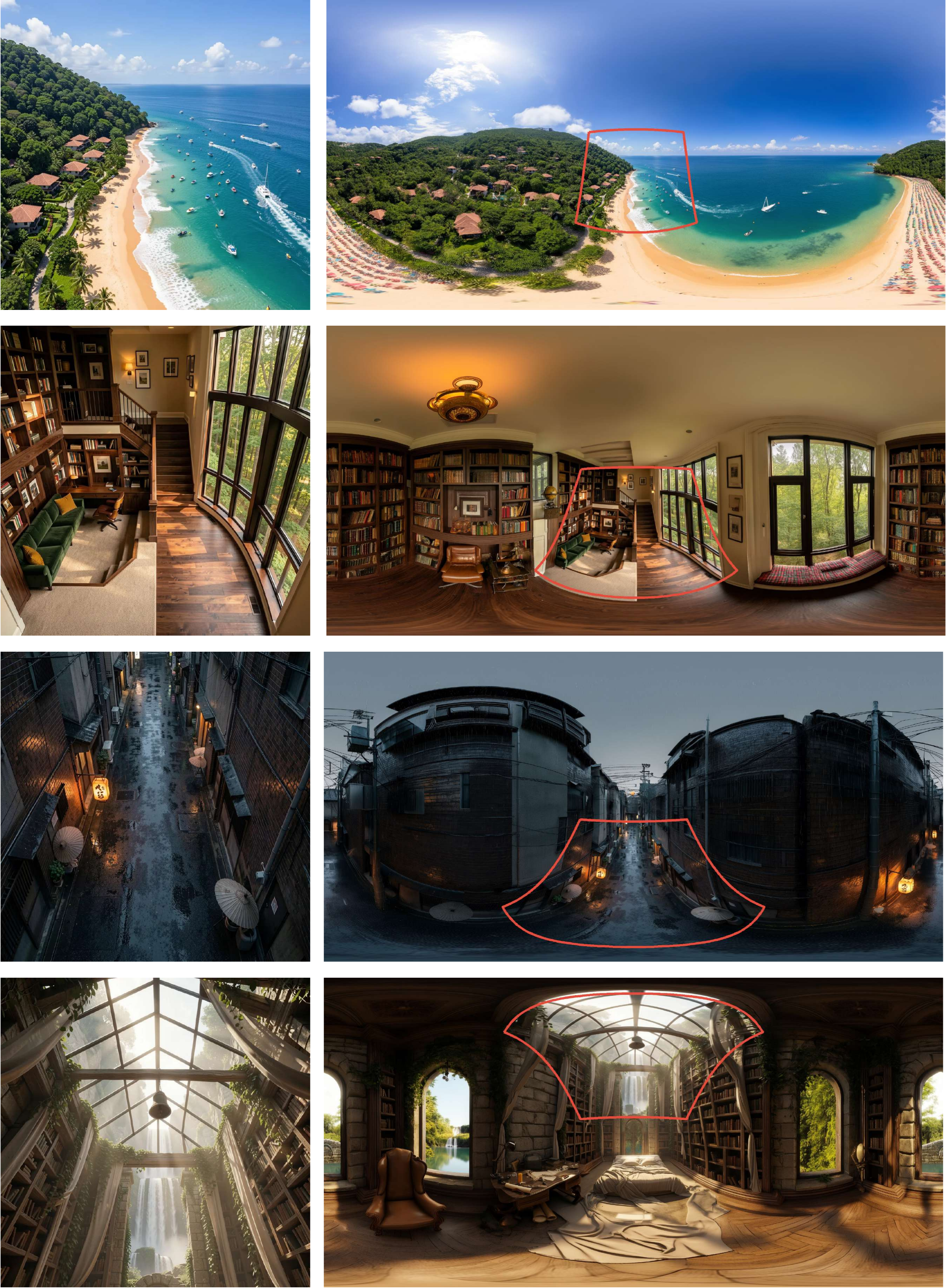}
\caption{\textbf{Panorama generation from diverse input viewing directions.} Each row shows a perspective input (left) and the generated ERP panorama (right). Our method accommodates downward- and upward-looking inputs without camera calibration or explicit input projection. Red outlines indicate the corresponding input regions in the generated panoramas.}
\label{fig:arbitrary-view-results}
\end{figure*}

\Cref{fig:arbitrary-view-results} demonstrates the versatility of our geometry-free formulation across perspective inputs with varying FoV and viewing directions, including upward- and downward-looking images. Rather than relying on camera calibration or explicitly projecting the input onto a predefined panoramic coordinate system, \methodname learns from data to infer the input view's placement in the panorama and synthesize its surrounding content accordingly.
Red outlines mark the regions in the generated ERPs that correspond to the input views.

\subsection{Additional Qualitative Results}
\label{sec:additional-qualitative}

\begin{figure*}[p]
\centering
\includegraphics[width=\textwidth]{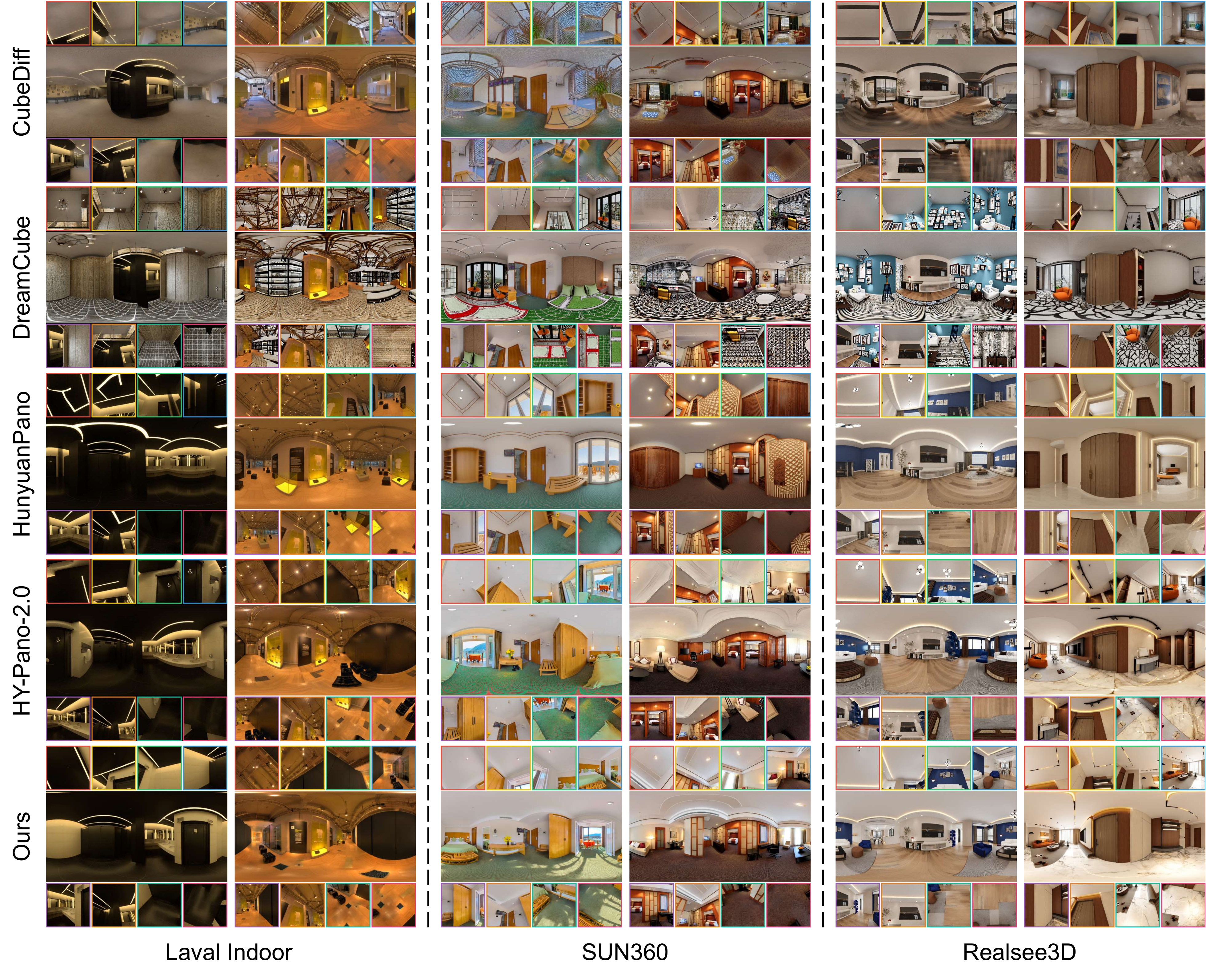}
\caption{\textbf{Additional qualitative comparisons} on Laval-Indoor, SUN360,
and RealSee3D. Results are shown for \methodname, HY-Pano-2.0, HunyuanPano,
DreamCube, and CubeDiff.}
\label{fig:additional-qualitative}
\end{figure*}

\begin{figure*}[p]
\centering
\includegraphics[width=\textwidth]{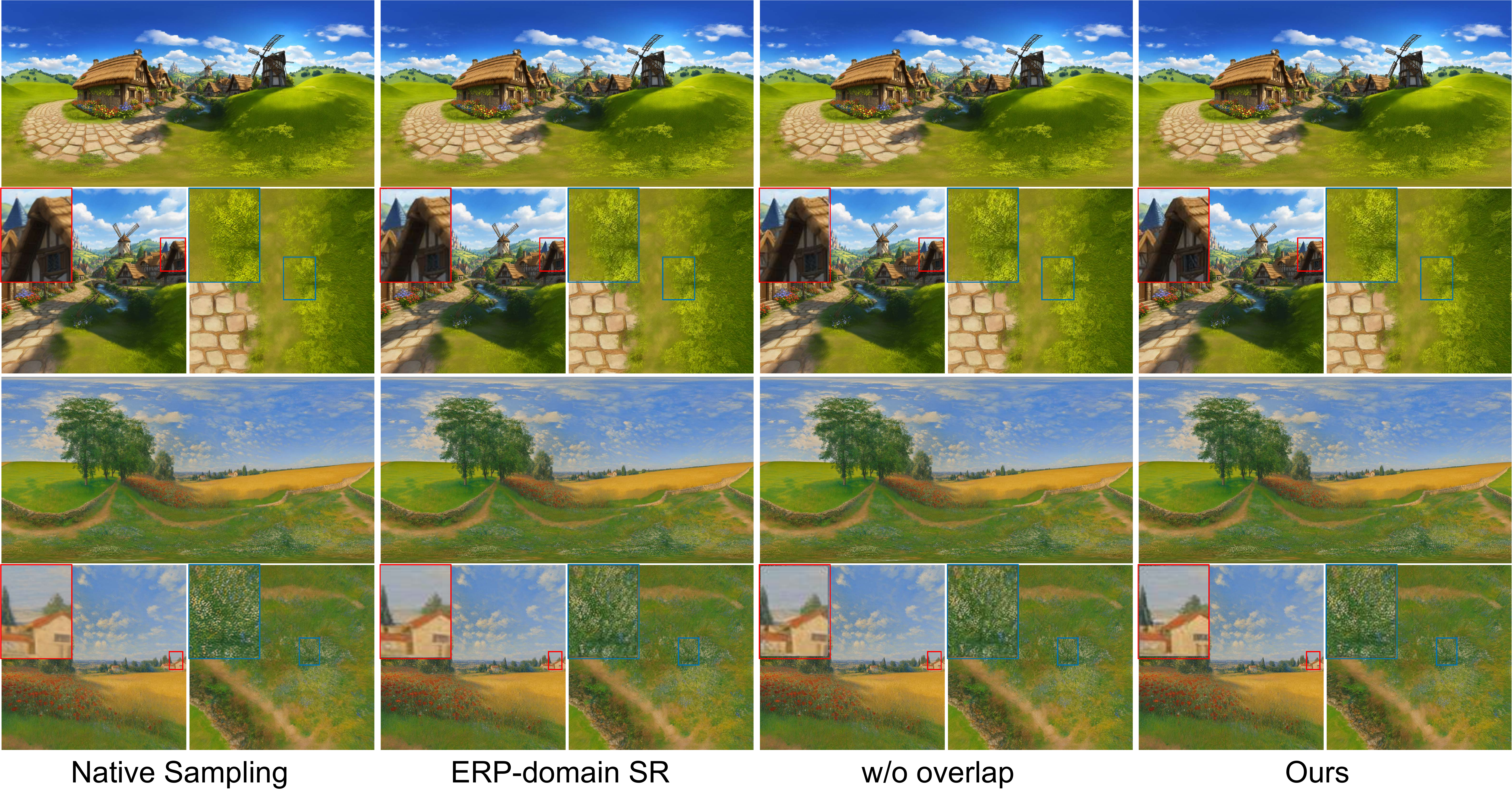}
\caption{\textbf{Additional qualitative results for high-resolution synthesis.}
From left to right: Native Sampling, ERP-domain SR, perspective-space SR without
overlap fusion, and the complete super-resolution pipeline. Enlarged crops compare
local detail recovery and cross-view consistency.}
\label{fig:additional-sr}
\end{figure*}

\begin{figure*}[p]
\centering
\includegraphics[width=0.95\textwidth]{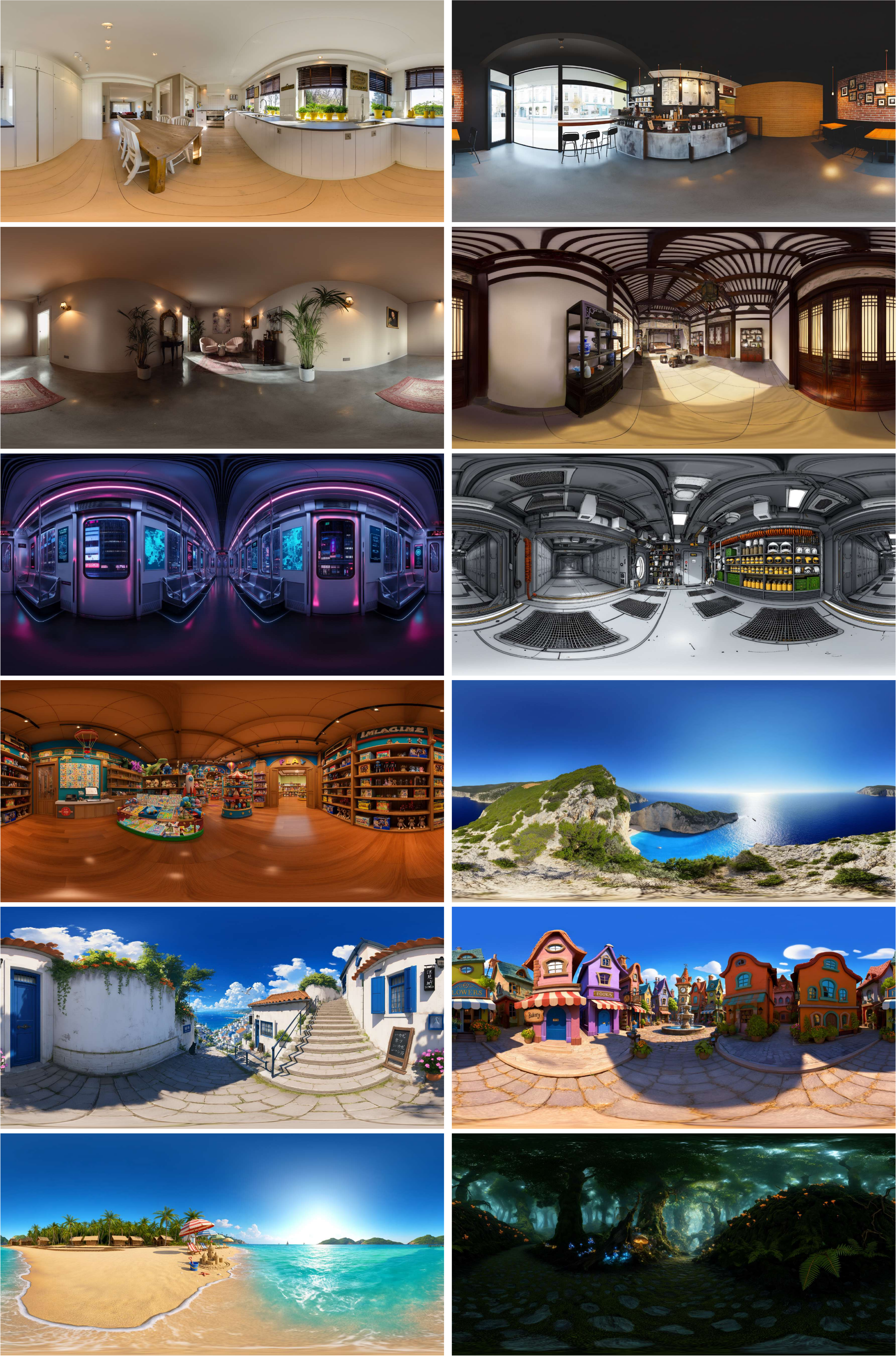}
\caption{\textbf{More qualitative results from \methodname.} We show diverse
indoor and outdoor scenes with realistic and stylized appearances. The generated
panoramas preserve the conditioning content and visual style while maintaining
coherent structures throughout the complete $360^\circ$ FoV.}
\label{fig:more-panorama-results}
\end{figure*}

\Cref{fig:additional-qualitative}, \Cref{fig:additional-sr}, and \Cref{fig:more-panorama-results} provide additional qualitative results.
\Cref{fig:additional-qualitative} includes comparisons on Laval-Indoor, SUN360, and RealSee3D across diverse indoor and outdoor scenes. \Cref{fig:additional-sr} shows high-resolution synthesis results obtained with different super-resolution strategies, including native sampling, ERP-domain super-resolution, perspective-space enhancement without overlap fusion, and our complete pipeline.
\Cref{fig:more-panorama-results} presents further generations spanning realistic, cinematic, stylized, and illustrative scenes.
Together, these examples provide a broader qualitative view of panorama generation across input viewing directions, cross-view consistency, and high-resolution detail synthesis.

\end{document}